\newif\ifarxiv 
\newif\ifcell
\arxivtrue  

\def \faketTitle {FakET: Simulating Cryo-Electron Tomograms with Neural Style Transfer}
\def \faketAffilMDS {Mathematical Data Science (MDS), Faculty of~Mathematics, University of~Vienna, Vienna, Austria} 
\def \faketAffilIMP {Haselbach Lab, Research Institute of~Molecular Pathology (IMP), Vienna, Austria}
\def \faketAffilRNDS {Research Network Data Science, University of~Vienna, Vienna, Austria} 
\def \faketAffilBDA {Brain Diseases Analysis Laboratory (BDALab), Brno University of~Technology, Brno, Czech Republic}
\def \faketAffilJRI {Johann Radon Institute for Computational and Applied Mathematics, Austrian Academy of~Sciences, Linz, Austria}
\def \faketAffilIST {Institute of Science and Technology Austria (ISTA), Klosterneuburg, Austria}
\def \faketKeywords {Machine Learning, Deep Learning, CryoET, Cryo-Electron Tomography, Neural Style Transfer, CryoEM, Deep Finder, Synthetic Data Generation, Pre-training, Domain Adaptation, Classification, Localization}
\def \faketContrib {Conceptualization, P.H. and D.H.; Methodology, Software, Validation, Formal Analysis, Investigation, P.H. and L.H.; Data Curation, and Visualization, P.H.; Resources, P.G.; Writing - Original Draft, P.H., P.G., and D.H.; Writing - Review \& Editing, P.H. and D.H.; Supervision, P.G. and D.H.; Funding Acquisition, P.G. and D.H.}
\def \acknowledgement {The IMP and D.H. are generously funded by Boehringer Ingelheim. We thank Julius Berner from the~Mathematical Data Science group @ UniVie, Ilja Gubins and Marten Chaillet from the~SHREC team, and the~members of~the~Haselbach lab for helpful discussions.}
\def \competing {The authors declare no competing interests.}
\def \contacts {P. Harar $<$pavol.harar@ista.ac.at$>$ and D. Haselbach  $<$david.haselbach@imp.ac.at$>$}

\documentclass[twocolumn,10pt,a4paper]{article}
\usepackage{caption}
\usepackage[table]{xcolor}
\usepackage{subfig}
\usepackage{authblk}
\usepackage{float}

\ifarxiv
\usepackage[round]{natbib}  
\fi

\ifcell
\usepackage[%
autocite=superscript,  
backend=biber,
bibstyle=apa,  
citestyle=numeric-comp,  
sorting=none,  
sortcites=true,
block=none
]{biblatex}

\makeatletter
\RequireBibliographyStyle{numeric}
\makeatother

\renewcommand{\cite}[1]{
    \autocite{#1}}
  
\fi

\ifarxiv 
\usepackage{fancyhdr} 

\fancypagestyle{bannerAccepted}{
    \setlength{\headheight}{17.55002pt}
    \addtolength{\topmargin}{-6.55002pt}
    \fancyhf{}  
    \fancyhead[C]{\scriptsize{\color{cyan} \textbf{ACCEPTED FOR PUBLICATION IN CELL STRUCTURE - DOI: \href{https://www.cell.com/structure/fulltext/S0969-2126(25)00020-6}{10.1016/j.str.2025.01.020}}} \\
    ~
    }
    \fancyfoot[C]{\thepage}  
}
\fi

\ifcell
\usepackage{fancyhdr} 
\fancypagestyle{customfooter}{
  \fancyhf{} 
  \fancyfoot[C]{Supplementary Information | \faketTitle} 
}
\fi

\ifcell
\renewcommand{\abstractname}{Summary}  
\fi

\renewenvironment{abstract}
  {
  \begin{center}
  \bfseries \hl{\abstractname}\vspace{-.5em}\vspace{0pt}
  \end{center}

  \small\bfseries
  \list{}{
    \setlength{\leftmargin}{0.3cm}
    \setlength{\rightmargin}{\leftmargin}%
  }%
  \item\relax}
 {\endlist}

\providecommand{\keywords}[1]{
\vspace{-2mm} ~\\
{\noindent\small\textbf{\textit{Keywords---}} #1} %
}

\providecommand{\contrib}[1]{
\vspace{-2mm} ~\\
{\noindent\small\textbf{\textit{Author Contribution---}} #1} %
}

\providecommand{\declaration}[1]{
\vspace{-2mm} ~\\
{\noindent\small\textbf{\textit{Declaration of Interests---}} #1} %
}

\providecommand{\ack}[1]{
\vspace{-2mm} ~\\
{\noindent\small\textbf{\textit{Acknowledgement---}} #1}}

\providecommand{\corresponding}[1]{
\vspace{-2mm} ~\\
{\noindent\small\textbf{\textit{Correspondence to---}} #1}}

\usepackage{microtype}
\usepackage{graphicx}
\usepackage{booktabs}
\usepackage{amsmath}
\usepackage{amssymb}
\usepackage{mathtools}
\usepackage{amsthm}
\usepackage{orcidlink}
\usepackage{hyperref}
\usepackage[capitalize,noabbrev]{cleveref}
\usepackage{doi}
\usepackage{xurl}
\usepackage{bm}  
\usepackage{balance}  
\usepackage{placeins}
\usepackage{bbding}  

\usepackage{multirow}

\newcommand\hl[1]{{#1}} 

\newcommand\hlb[1]{{#1}} 

\usepackage{etoolbox}
\makeatletter
\patchcmd\@addmarginpar{\hb@xt@}{\pdfrunninglinkoff\hb@xt@}{}{\fail}
\apptocmd\@addmarginpar{\pdfrunninglinkon}{}{\fail}
\makeatother

\begin{document}

\title{\faketTitle}

\ifarxiv
\author[1,2,3,4,5]{\orcidlink{0000-0001-5206-1794}Pavol Harar}
\author[6]{\orcidlink{0000-0003-3402-6420}Lukas Herrmann}
\author[1,3,6]{\orcidlink{0000-0001-9205-0969}Philipp Grohs}
\author[2]{\orcidlink{0000-0002-5276-5633}David Haselbach}
\fi

\renewcommand\Affilfont{\footnotesize} 
\affil[1]{\faketAffilMDS} 
\affil[2]{\faketAffilIMP}
\affil[3]{\faketAffilRNDS}
\affil[4]{\faketAffilBDA}
\affil[5]{\faketAffilIST}
\affil[6]{\faketAffilJRI}

\date{}

\ifarxiv
\renewcommand{\cite}[1]{
    \citep{#1}}
\fi

\twocolumn
\ifarxiv
\makeatletter
\def\@maketitle{%
  \newpage
  \begin{center}%
  \let \footnote \thanks
    {\LARGE \@title \par}%
    \vskip 1.5em%
    {\large
      \lineskip .5em%
      \begin{tabular}[t]{c}%
        \@author
      \end{tabular}\par}%
    \vskip 1em%
    {\large \@date}%
  \end{center}%
  \par
  \vskip 1.5em}
\makeatother
\fi
\maketitle

\begin{abstract}

\hl{
In cryo-electron microscopy, accurate particle localization and classification are imperative. Recent deep learning solutions, though successful, require extensive training data sets. The protracted generation time of physics-based models, often employed to produce these data sets, limits their broad applicability. We introduce FakET, a~method based on Neural Style Transfer, capable of simulating the forward operator of any cryo transmission electron microscope. It can be used to adapt a~synthetic training data set according to reference data producing high-quality simulated micrographs or tilt-series. To assess the quality of our generated data, we used it to train a~state-of-the-art localization and classification architecture and compared its performance with a~counterpart trained on benchmark data. Remarkably, our technique matches the performance, boosts data generation speed~$\bm{750\times}$, uses $\bm{33\times}$\,less memory, and scales well to typical transmission electron microscope detector sizes. It leverages GPU acceleration and parallel processing. The source code is available at \href{https://github.com/paloha/faket/}{https://github.com/paloha/faket/}.
}
\end{abstract}

\ifarxiv
\vfill

\ack{\acknowledgement}

\contrib{\hl{\faketContrib}}

\declaration{\hl{\competing}}

\corresponding{\contacts}
\fi

\ifarxiv
\thispagestyle{bannerAccepted} 
\fi

\ifarxiv
\begin{center}
\bfseries \hl{Graphical abstract}\vspace{0.65em}\vspace{0.55em} \\
\centerline{\includegraphics[width=\columnwidth]{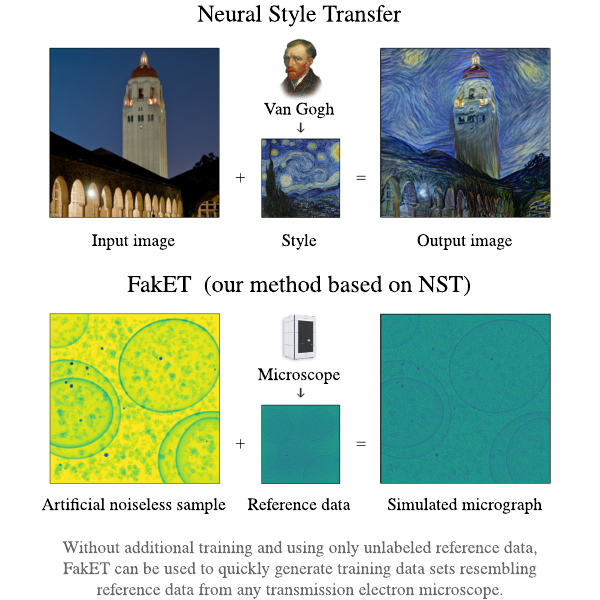}}
\end{center}
\fi

\section{Introduction}
\label{sec:introduction}

Recent developments in~cryo-electron tomography (cryoET) allow to~obtain high resolution representations of~macromolecular complexes in~their native cellular environment showing molecular interactions that are hardly accessible with other methods\cite{Turk2020}. 
\ifarxiv
\begin{table*}[!ht]
\vskip 0.15in
\begin{center}
\begin{small}
\begin{sc}
\begin{tabular}{llllll}
\toprule
Model       & Training data             & Data cost                                      & Localiz.\@ F1   & Classification F1 \\
\midrule
DeepFinder  & {\sc benchmark}           & $\approx150\,h$ (3$\times$CPU, 114\,GB RAM)    & 0.815             & 0.581 ($100\,\%$)\\
\ifarxiv\rowcolor{lightgray!40}\fi
DeepFinder  & {\sc faket} + fine-tuning & {\sc faket} cost + manual labeling             & 0.821             & 0.565 ($97\,\%$) \\
\ifarxiv\rowcolor{lightgray!40}\fi
DeepFinder  & {\sc faket}               & $\approx12\,min$ (1$\times$GPU, 40\,GB VRAM)     & 0.800             & 0.533 ($92\,\%$) \\
\ifarxiv\rowcolor{lightgray!40}\fi
            &                           & \hlb{or $\approx80\,min$ (8$\times$CPU, $\approx6.5$\,GB RAM)}     &               &  \\
DeepFinder  & {\sc baseline}            & $\approx20\,s$ (1$\times$CPU, 1\,GB RAM)       & 0.813             & 0.441 \\
TM-F        &                           &                                                & 0.576             & 0.446 \\
TM          &                           &                                                & 0.372             & 0.470 \\
\bottomrule
\end{tabular}
\end{sc}
\end{small}
\end{center}
\vskip -0.2in
\ifcell\captionsetup{labelfont=bf}\fi
\def \tableTitle {\hl{Evaluation of simulated data quality by examining}~DeepFinder's performance on~localization and classification tasks as a~function of~training data.}
\caption{
\ifcell\textbf{\tableTitle}\else{\tableTitle}\fi
~The~performance is evaluated on~the~same test tomogram. The~results are shown in~context with the~performance of~standard template matching algorithms (denoted TM and TM-F) reported in\cite{shrec2021} for the~same testing data. Performance is measured using the~F1 score for localization and the~F1 macro score for classification. Each of~the~DF results is an average of~the~test performances measured at the~best epoch (based on~validation) over 6 different random seeds. On the~challenging classification task, DF trained using {\sc faket} data simulated by our proposed method reaches 92\,\% of~the~performance of~DF trained using {\sc benchmark} data. It even reaches 97\,\% when fine-tuned using a~portion of~{\sc benchmark} data. All this for a~fraction of~computational cost and without the~need for a~configuration protocol of~the~original simulation parameters used for crating {\sc benchmark}. \hl{The cost of data is given for the whole data set comprising 10 tilt-series of shape ($61\times1024\times1024$), i.e. generating one such tilt-series with {\sc faket} takes $\approx70$ seconds assuming a~GPU is available. For comparison, a~usual-sized tilt-series of shape ($61\times3500\times3500$), that is $\approx12\times$\,larger in comparison to {\sc benchmark}, can be simulated under 10 minutes. \hlb{For a comprehensive time and memory consumption profiling of {\sc faket}, see \Cref{appendix:profiling}.}}}
\label{table:result}
\end{table*}
\fi
In cryoET, the~imaged sample is in~most cases a~100-200\,nm thick slice of~a~frozen cell. From this slice, projection images are taken in~a~transmission electron microscope~(TEM) from different rotation (tilt) angles. An artifact free reconstruction would require measurements using tilt angles, that would complete the~half circle. However, this is not feasible, due to~limitations of~the~specimen holder, and only a~range of~140$^{\circ}$ can be recorded. The~missing tilt images later on~result in~a~so-called missing-wedge in~the~3D reconstruction (cryo-electron tomogram). In addition, the~electron beam severely damages the~sample during imaging, so only a~low electron dose can be used to~image a~biological specimen. The~low dose in~combination with the~presence of~ice in~the~sample results in~the~acquired data being very noisy. Consequently, the~identification of~molecules within these reconstructions is a~daunting task. Particle identification is however necessary as the~particles need to~be classified and averaged to~determine high resolution structures. While cryoET has led to~a~large number of~breakthroughs, providing hitherto unseen detail in~the~molecular architecture of~cells\cite{zimmerli2021,oreilly2020, mahamid2016}, the~aforementioned challenges still hinder the~widespread use of~cryoET in~the~larger cell biology and structural biology community. In this context, the~development of~new reliable software tools is of~paramount importance, which is however obstructed by the~lack of~sufficient accessible and annotated data to~develop the~software tools~on.

\ifcell\else
\subsection{SHREC simulator} \fi
\label{subsec:shrecsimulator}
To overcome the~problem with the~lack of~data, in~2019, the~annual \emph{SHREC -- 3D Shape Retrieval Contest} included a~new track titled \emph{Classification in~Cryo-Electron Tomograms}. The~organizers of~this track proposed a~task of~localization and classification of~biological particles in~cryo-electron tomograms. In the~following years, experts from 3D object retrieval and 3D electron microscopy communities were invited to~participate in~the~challenge. In order to~ensure fair evaluation and comparable results across the~submissions, the~organizers created a~data set of~ten physics-based cryo-electron tomogram simulations (9 train \& 1 test tomogram, see~\Cref{fig:projshrec}) for the~contestants to~train and evaluate their methods on. Each year, the~results of~the~contesting methods~were presented and compared\cite{shrec2019, shrec2020, shrec2021}.

Unfortunately, simulating the~tomographic data using SHREC is computationally very expensive. For a~set of~10 small tilt-series (61 tilts of~size $1024 \times 1024$) it took approx.\,\hl{4}50~CPU hours of~computation on~a~node with \emph{$2\,\times$\,Intel Xeon E5-2630 v4} CPUs. The~implementation is able to~utilize only 3 CPU cores per job in~parallel and each job needs 114\,GB of~RAM. The~memory consumption is also a~reason why utilizing GPUs and simulating tilt-series of~common sizes is not yet feasible (personal communication with the~authors Gubins,~I., and Chaillet,~M.). This limits the~scope of~its applicability, mainly in~data-hungry applications such as deep learning. Moreover, at the~time of~writing of~this article, the~source code of~the~simulator is not publicly available \hl{prohibiting more detailed comparisons, e.g. in terms of FLOPs}.

\ifcell\else
\subsection{Our contribution} \fi
\label{subsec:contrib}


In this paper, we propose {\sc faket}, a~\hl{fast and scalable data-driven} method for simulating the~forward operator of~\hl{any}~cryo transmission electron microscope \hl{with the aim} to~generate synthetic \hl{micrographs or} tilt-series.
It was created, among other reasons, to~generate \hl{fully-labeled real-like} data for \hl{training} deep neural networks to~solve tasks such as particle localization and (much more challenging) particle classification. 
Our method combines a noiseless simulated sample, see~\Cref{fig:projnoiseless}, with additive noise, see~\Cref{fig:projbaseline}, and neural style transfer~(NST) technique based on\cite{Gatys_2016_CVPR} \hl{to capture, using only unlabeled reference data, the structure of the complex noise introduced by TEM, see~\Cref{fig:projfaket}.}
\hl{To carry out the NST, {\sc faket} utilizes a~pretrained model, eliminating the need for users to train it themselves.} 
\hl{That means users can use {\sc faket} to simulate data from any TEM and under any configuration, assuming they posses unlabeled TEM data that could serve as a~style reference.}
\hl{{\sc faket} delivers data of quality nearly identical}%
\ifarxiv
\footnote{\hl{In terms of their practical utility for subsequent tasks, rather than exact manifestation of the physical interactions between the sample and the transmission electron microscope.}}
\fi
\hl{to the reference, at a~cost only slightly higher than simply adding Gaussian noise to a~noiseless synthetic sample}%
\ifarxiv
\footnote{\hl{Created using existing models of biological macromolecular structures, represented as Coulomb density volume.}}.
\fi
\hl{This makes it a~practical, effective, and efficient choice for simulation.}

\hl{For the purpose of evaluating our method, in the experiments presented within this article, we employed {\sc faket} to mimic the~behavior of~the~physics-based TEM simulator SHREC}%
\ifarxiv
\footnote{\hl{Due to the availability of ground truth.}}.
\fi
\hlb{On top} \hl{of evaluating the quality of our~simulated data using standard image metrics} \hlb{(available in the supplement)} \hl{that are of limited value in this context, we evaluated it directly on practically-relevant} downstream tasks by training DeepFinder~(DF)\cite{Deepfinder} -- a~neural network specifically tailored to~the~task of~particle localization and classification, see also
\ifarxiv \Cref{subsec:deepfinder}.\fi

In contrast to~SHREC, our method accelerates the~data generation process by a~factor of~750 while using 33 times less memory (see~\Cref{table:result}). It also does not require any calibration protocol as other simulators, see
\ifarxiv \Cref{subsec:simulators}. \fi
\hl{The NST model} does not need to~be retrained to~be used on~new data, nor does it require labeled reference data. \hl{{\sc faket}} therefore has the~potential to~save experts countless hours of~manual work in~labeling their data sets.%
\ifarxiv
\footnote{Depending on~the~imaged sample, an expert may spend several hours of~manual work per tilt-series per particle labeling the~data. At the~same time, certain smaller particles cannot be visually found at all, thus fully labeled tomograms do not really exist.}
\fi
Moreover, \hl{\sc faket} is capable of~simulating large tilt-series, which are common in~experimental environments (about $~12\times$ larger than SHREC projections). For example, we generated a~$61\times3500\times3500$ tilt-series on~a~single \emph{NVIDIA A100 40GB SXM4} GPU in~less than 10 minutes.%
\ifarxiv
\footnote{\hl{To achieve the reported speeds, a~minimum of one GPU is recommended; \hlb{CPU-only simulations are feasible and can be a valid option in specific situations, albeit slower.}}}
\fi
This advance makes it possible to~train particle localization and classification networks from scratch, or to~pre-train networks that are later fine-tuned using manually labeled experimental data, see~\Cref{appendix:finetuning}. 
On top of~that, our method is open source and our experiments are reproducible. 

Additionally, we include comparisons of~the~{\sc benchmark} and the~proposed {\sc faket} method with a~simple {\sc baseline} method based on~additive Gaussian noise, see~\Cref{fig:projbaseline},
\hl{aiming at providing comparative results in full practical range. The costly SHREC method defines an upper-boundary of performance, while the cheap addition of Gaussian noise establishes a~lower boundary. This configuration enables us to position our method along this spectrum, showcasing {\sc faket}'s capability to offer results close to the upper-boundary while maintaining computational efficiency of a~much simpler method. We hope our comparisons will be useful for practitioners who need to~decide which method fits into their computational budget and for those seeking insightful understanding of the inherent trade-offs.}
The~contribution of~our method is further supported by an ablation study presented in~\Cref{appendix:additionalexps}, where we also offer insight into the~potential limits of~DF on~the~studied data by conducting experiments using completely noiseless simulated tomograms.

\ifcell\else
\subsection{Related work} \fi
\label{subsec:relatedworks}
A~similar idea in~X-ray-based computed tomography angiography (CTA) was investigated in\cite{seemann2020data}. The~authors focused on~solving a~lumen segmentation task. However, due to~the~very different nature of~the~samples imaged in~CTA%
\ifarxiv
\footnote{The objects of~interest in~CTA are on~average two orders of~magnitude larger in~relation to~the~size of~the~tomogram than those in~cryoET, where the~particles often span only tens of~voxels.}
\fi
, it was not clear whether a~similar NST based framework could be successfully applied in~cryoET to~image nanoscale particles. In addition, the~article does not provide source code and does not document the~experiments in~enough detail for us to~be able to~reproduce the~results or adapt the~method to~the~cryoET domain. 

\ifcell\else
\subsection{Other simulators} \fi
\label{subsec:simulators}
The value of~simulated data in~cryoEM is well recognized and micrograph simulation has been attempted several times. In first approximation a~single cryoEM micrograph is the~projection of~a~3D object, convolved with the~electron microscope's point spread function. Additionally, the~overall process comprises several sources of~noise coming from the~nature of~the~sample, the~microscope, and the~imaging procedure which are hard to~model accurately. For a~better overview of~the~attempts, it is necessary to~mention TEM simulators that were developed in~the~past decades to~simulate micrographs in~cryo conditions. 

Earlier works provided fast simplistic models mostly based on~additive white Gaussian noise%
\ifarxiv
\footnote{This motivated our choice of~baseline.}
\fi
or coloured noise\cite{scheible2021tomosim}. Other works originated from the~insights into the~physics of~TEM image formation and advanced the~simulations by modeling various sources of~noise, e.g.~an improvement in~modeling the~structural noise was proposed in~\emph{TEM~Simulator}~(C)\cite{rullgaard2011simulation}. Another improvement was done in~\emph{InSilicoTEM} (Matlab) presented in\cite{vulovic2013image} by implementing the~multislice method originally proposed in\cite{cowley1957scattering}. This makes it the~next most relevant simulator related to~our work after SHREC which is, as many others, also based on~the~same multislice method%
\ifarxiv
.\footnote{This motivated our choice of~benchmark.}
\fi
To the~best of~our knowledge, the~most recent improvement presented in\cite{himes2021cisTEM} is distributed as a~part of~\emph{cisTEM} package~(C++). It introduced frozen plasmon method to~explicitly model spatially variable inelastic scattering processes in~cryo-electron microscopy. The~aforementioned works, based on~the~same multislice method, however, suffer from a~heavy computational burden or are limited to~simulations of~a~single molecular complex. Despite these limitations, they represent a~set of~precise advanced physics-based simulators and as such also provide a~comprehensive literature survey referring the~reader to~a~body of~detailed resources about modeling the~image formation in~transmission electron microscopy. 

In material sciences, GPU accelerated simulators such as \emph{MULTEM}~(C++, CUDA)\cite{lobato2015multem, lobato2016multemII}, \emph{abTEM}~(Python)\cite{madsen2020abtem}, or \emph{Prismatic}~(C++, CUDA)\cite{pryor2017prismatic} have emerged. However, as reviewed in\cite{kirkland2020advanced}, advanced TEM simulators used in~material sciences require atomic models of~the~background and the~entire specimen. Moreover, in~cryoEM, the~sample and its interaction with the~electron beam is fundamentally different. While in~material science samples are often only a~few layers of~atoms thick and the~imaged atoms have strong interaction with the~electron beam, cryoEM samples are thousands of~atom layers thick, are less ordered, and only weakly interact with the~electron beam. Additionally, cryoEM samples can only withstand little radiation before complete destruction resulting in~much lower signal to~noise ratios.
Due to~these differences, such simulators are of~limited use for macromolecular biological specimen simulations that we experiment with in~this article. Finally, in~contrary to~our approach, all aforementioned methods require precise calibration protocols for setting the~values of~simulation parameters.

\ifcell\else
\subsection{Structure of~this article} \fi
\label{subsec:structure}
To build our proposed method, presented in
\ifarxiv \Cref{sec:methods} \fi
(referred to~as {\sc faket}), we used a~data-driven approach (i.e. no calibration protocol is needed). More information about the~data we used is in
\ifarxiv \Cref{sec:data}. \fi
With our method, we managed to~successfully approximate the~input-output behavior, on~par with the~SHREC simulator (referred to~as {\sc benchmark}), but for a~fraction of~its computational cost. To experimentally prove the~need for our method, we also compared it to~a~simple addition of~Gaussian noise (referred to~as {\sc baseline}) that was naturally very fast to~compute, but did not lead to~comparable results on~the~downstream tasks.

To evaluate the~methods, we used the~simulated data by each of~them as a~training set for the~DeepFinder neural network (more in
\ifarxiv \Cref{subsec:deepfinder}\fi
). The~network was trained to~solve two evaluation tasks proposed in~SHREC challenge. The~final comparison was done by observing the~models' performances on~the~test tomogram chosen by SHREC. Detailed description of~the~evaluation is in
\ifarxiv \Cref{sec:evaluation}. \fi
More on~experiments and results in
\ifarxiv \Cref{sec:results}. \fi
Procedures of each method are compared in~\Cref{fig:diagram}.
This paper is accompanied with source code, the~results are fully-reproducible, and the~full experiment results are available in~the~repository -- \href{https://github.com/paloha/faket/}{https://github.com/paloha/faket/}. 

Side-by-side comparison of~all projections used in~this article is available in~\Cref{appendix:sidebyside}. The~process of~re-creating the~{\sc benchmark} data is detailed in~\Cref{appendix:benchmark}. Fine-tuning experiments that further improved the~DF performance are described in~\Cref{appendix:finetuning}. Additional experiments, NST ablation study, and investigation of~DF limits are presented in~\Cref{appendix:additionalexps}. Further details on~the~performance of~studied models on~the~particle classification task are provided in~\Cref{appendix:confmats}.

\ifcell\else
\subsection{Key findings} \fi
\hl{
     {\sc faket} successfully utilizes NST to simulate cryo-electron micrographs or tilt-series of common sizes.
     •~{\sc faket} produces realistic TEM simulations of quality nearly identical to the unlabeled reference data.
     •~Cost of {\sc faket} simulation is only slightly higher than simply adding Gaussian noise while significantly lower than that of complex physics-based simulators.
     \hlb{•~Apart from unlabeled reference data, {\sc faket} does not require further training or calibration.}
     •~{\sc faket} is a~practical and efficient tool for simulating fully-labeled training data for deep learning.
}


\def \figProjshrecTitle {Simulated projection taken from SHREC 2021 data set.}
\def \figProjshrecCaption {Axes x and y correspond to~width and height of~the~imaged grandmodel. Colorbar denotes simulated intensities in~arbitrary units. See~\Cref{appendix:sidebyside} for side-by-side comparison with other projections.}
\def \figProjnoiselessTitle {Noiseless projection used to~create the~input to~our proposed method.}
\def \figProjnoiselessCaption {Axes x and y correspond to~width and height of~the~imaged grandmodel. Colorbar denotes intensities measured using Radon transform and negated such that particles have lower intensities than the~background, as it is in~the~case of~TEM which measures attenuation of~electron beams. The~particles are not embedded in~any solvent (as if they were in~vaccuum instead of~being embedded in~ice), therefore the~background appears much brighter than in~the~simulated projections. See~\Cref{appendix:sidebyside} for side-by-side comparison with other projections.}

\ifarxiv
\section{Data}
\label{sec:data}
For the~purposes of~this paper, we used the~latest version of~the~SHREC data~set from the~year 2021. More specifically, we based the~presented methods on~full-resolution grandmodels (synthetic volumes containing randomly scattered particles) and simulated projections from the~\emph{SHREC\,2021 additional} data~set that was made publicly available%
\ifarxiv
\footnote{Download v2.0: \href{https://www.shrec.net/cryo-et/}{shrec.net/cryo-et}, \doi{10.34894/XRTJMA}.}.
\fi
\ifcell
~(more info at \href{https://www.shrec.net/cryo-et/}{shrec.net/cryo-et}, download version 2.0 from \doi{10.34894/XRTJMA}).
\fi
An example visualization of~a~simulated projection using the~SHREC simulator is depicted in~\Cref{fig:projshrec}. We use the~unchanged simulated projections as a~benchmark which we try to~approximate with our proposed method.

The steps to~create the~simulations were described in\cite{shrec2021}. However, at the~time of~writing of~this paper, the~implementation is not publicly available and from the~description of~the~steps it is clear that the~method is, and rightfully so, very elaborate, and not at all elementary to~reimplement (also due to~various steps being under-documented, see~\Cref{appendix:benchmark}). 

On the~following lines, we briefly summarize the~steps needed to~create the~SHREC data set. The~authors first constructed 3D ground-truth specimens (grandmodels) along with annotations. Each specimen contained uniformly distributed and rotated protein instances (represented as Coulomb density) of~varying size and structure from the~Protein Data Bank (PDB)\cite{PDB}, as well as membranes and gold fiducials, which are commonly found in~tomograms. Then, they simulated a~layer of~amorphous ice into the~grandmodels before rotating them over 61 evenly-spaced tilt angles, ranging from -60$^{\circ}$ to~+60$^{\circ}$, in~order to~be projected. The~noiseless projections were produced using their own implementation of~a~TEM simulator based on~the~multislice approach presented in\cite{vulovic2013image}. Next, they sampled from a~Poisson distribution with a~specific electron dose to~obtain the~final electron counts (files labeled as \textit{projections}). After that, they randomly shifted the~projections to~model the~tilt misalignment and scaled the~amplitudes in~Fourier space using information about amplitudes from experimental images in order to~increase their similarity (files labeled as \textit{projections\_unbinned}). In the~end, they obtained the~final images (files labeled as \textit{reconstruction}) by reconstructing the~$2\times$~binned projections using weighted-backprojection algorithm from a~private version of~the~PyTom package\cite{pytom}.

We chose the~SHREC data set as it is established, well-executed, downloadable, and allows researchers to~compare their results with previous works without the~need to~reproduce all results from scratch. In this paper, we are not focusing on~solving the~proposed tasks, instead, we are using them as evaluation metrics to~measure the~performance of~our proposed method. However, the~availability of~the~data and descriptions of~the~methods made this research feasible.

\ifarxiv
\begin{figure}
\vskip 0.2in
\begin{center}
\centerline{\includegraphics[width=\columnwidth]{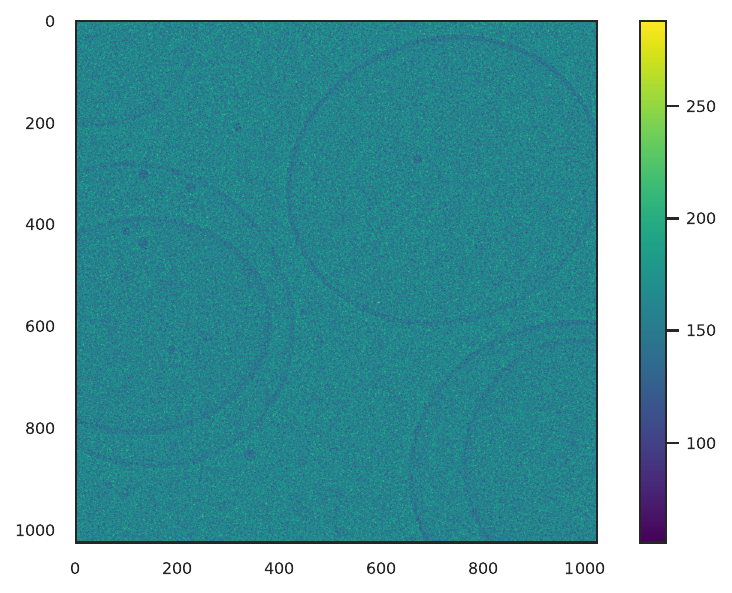}}
\caption{\figProjshrecTitle~\figProjshrecCaption}
\label{fig:projshrec}
\end{center}
\vskip -0.2in
\end{figure}
\fi

As stated earlier, we used the~full-resolution grandmodels from the~SHREC data set to~create our own noiseless projections using Radon transform, as implemented in~the~\emph{scikit-image} library for image processing in~Python\cite{scikit}. We computed our own noiseless projections for three primary reasons. Firstly, SHREC only supplied full-resolution noiseless projections embedded within the~simulated ice layer, a~feature we also aimed to~incorporate in~our surrogate. Secondly, we intended to~provide access to~all implementation steps within the~code-base. Lastly, it was necessary to~produce noiseless reconstructions that could later be used to~estimate the~performance boundaries of~DF on~this data set.
We also had to~omit the~usage of~the~reconstructions provided by SHREC because the~exact configuration of~all the~steps is not published and the~version of~the~PyTom software package used to~create this data is not public. Also, the~public version is not yet properly documented or straightforward to~use. Thus, the~only feasible option to~ensure comparable results across our experiments was to~create our own reconstructions from all the~newly created projections described in
\ifarxiv \Cref{sec:methods}. \fi
\ifcell \nameref{sec:methods}. \fi
Nevertheless, we put a~lot of~effort into matching our reconstructions with the~ones originally published in~SHREC. We created them using the~RadonTea Python package\cite{radontea} using a~custom filter described in~\Cref{appendix:benchmark}. For additional flexibility, we also implemented support for tomographic reconstructions using the~well-known IMOD package\cite{IMOD}. To ensure comparability also with previously published results, we used the~original SHREC \emph{model\_9} reconstruction as a~test tomogram for all presented methods as it was done in~all the~past challenges.

\ifarxiv
\begin{figure}[!t]
\vskip 0.2in
\begin{center}
\centerline{\includegraphics[width=\columnwidth]{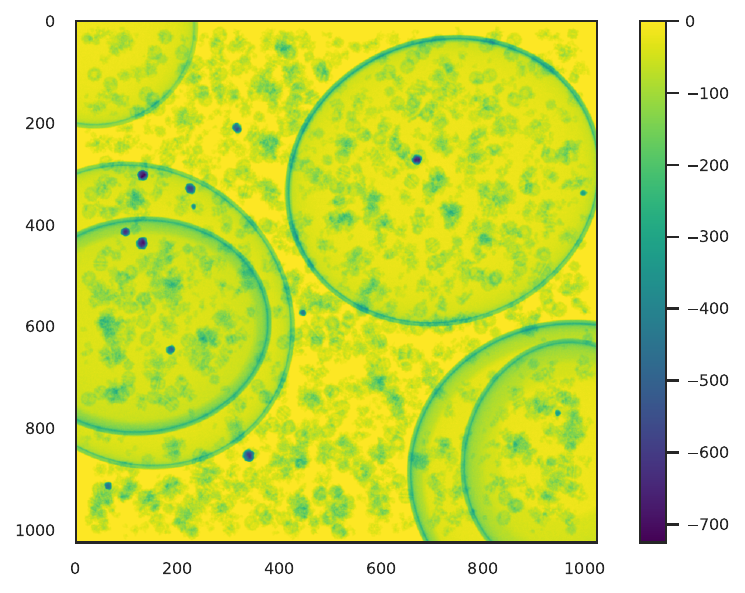}}
\caption{\figProjnoiselessTitle~\figProjnoiselessCaption}
\label{fig:projnoiseless}
\end{center}
\vskip -0.2in
\end{figure}
\fi
\else
\refstepcounter{figure}\label{fig:projshrec}
\refstepcounter{figure}\label{fig:projnoiseless}
\fi


\def \figProjbaselineTitle {{\sc baseline} projection created by adding Gaussian noise to the noiseless projection.}
\def \figProjbaselineCaption {Axes x and y correspond to~width and height of~the~imaged grandmodel. Colorbar denotes simulated intensities in~arbitrary units. Please note that this projection is not exactly the~same as in~\Cref{fig:projshrec} or in~\Cref{fig:projfaket}, see also the~explanation in~the~caption of~\Cref{fig:projfaket} and \Cref{appendix:sidebyside} for a~side-by-side comparison with other projections.}
\def \figProjfaketTitle {{\sc faket} projection output by our method.}
\def \figProjfaketCaption {Axes x and y correspond to~the~grandmodel's width and height respectively, and colorbar indicates simulated intensities in~arbitrary units. Though visually similar to~\Cref{fig:projbaseline}, there is a~subtle difference that has a~significant impact on~DF's performance. This similarity complicates comparison of~projections from various simulators using currently available image metrics. See~\Cref{appendix:sidebyside} for a~side-by-side comparison with other projections.}
\def \figDiagramTitle {Diagram of~steps to~simulate the~{\sc benchmark}, {\sc baseline}, {\sc faket}, and noiseless projections and reconstructions.}
\def \figDiagramCaption {Red arrows highlight steps to~reproduce SHREC data from which we use the~last tomogram for testing. All methods except SHREC were filtered using a~reverse-engineered filter (see~\Cref{appendix:benchmark}) because the~SHREC filtering step is under-documented. The~style projections never feature the~same contents as the~simulated ones (see~\Cref{appendix:sidebyside}). \hl{Grandmodels, noiseless artificial samples containing randomly scattered particles, were created using existing models of biological macromolecular structures, represented as Coulomb density volumes.}}

\ifarxiv
\section{Methods}
\label{sec:methods}
\ifarxiv
\begin{figure}[!t]
\vskip 0.2in
\begin{center}
\centerline{\includegraphics[width=\columnwidth]{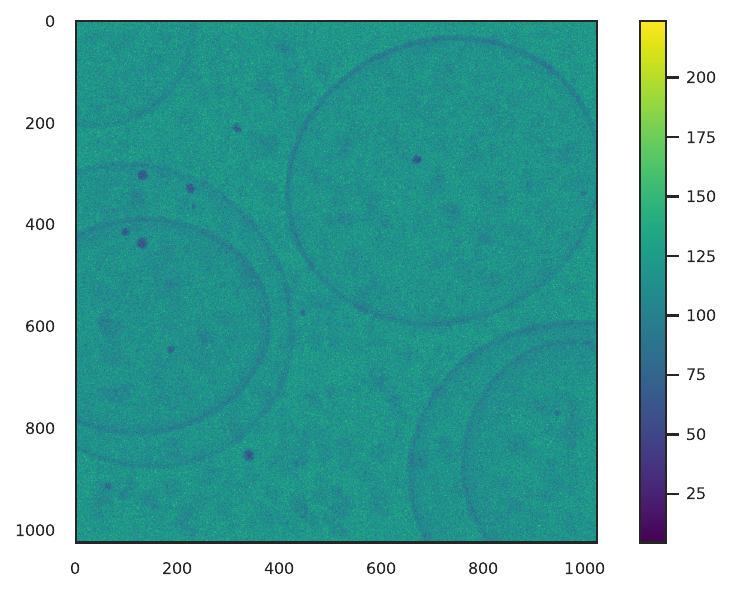}}
\caption{\figProjbaselineTitle~\figProjbaselineCaption}
\label{fig:projbaseline}
\end{center}
\vskip -0.2in
\end{figure}
\fi

The methods described in~this section represent surrogate models that mimic the~behaviour of~the~aforementioned SHREC physics-based simulator. The~methods are to~be applied in~the~projection space, i.e.~all of~them take noiseless (and ice-less) projections as inputs, shown in~\Cref{fig:projnoiseless}, and produce real-like looking projections by matching their appearance with a~target ``style'' projections. The~outputs are ready to~be reconstructed into final tomogram volumes using a~reconstruction algorithm of~choice. 
There are multiple reasons why we decided, and why it is reasonable, to~model the~simulator's behaviour in~projection space. Firstly, different reconstruction algorithms produce different types of~artifacts and are suitable in~different situations. Therefore, simulating only projections means the~models do not have to~mimic the~artifacts crated by the~reconstruction algorithm (e.g.~smearing due to~the~missing wedge). Moreover, this approach does not limit the~practitioners in~their choice of~the~reconstruction algorithm, and opens the~possibility for researchers to~also use the~methods in~research of~novel reconstruction algorithms. But most importantly, simulating in~projection space means the~models need to~process and produce an order-of-magnitude less data points, as compared to~simulating the~final reconstruction volumes. That means, simulating $\theta \times N^2$ data points instead of~$N^3$. In our case, $\theta = 61$ and $N = 1024$. In practice, $\theta$ stays approximately the~same, but $N$ is usually 3\,--\,4$\times$~larger. 

\ifcell \subsubsection*{Additive noise ({\sc baseline})} \else
\subsection{Additive noise ({\sc baseline})} \fi
\label{subsec:baseline}

Before moving on~to the~development of~more complicated methods, we wanted to~see how a~very simple method involving tilt-dependent scaling and the~addition of~Gaussian noise would approximate the~target projections. The~modality of~projections produced by this method is referred to~as {\sc baseline} projections in~the~whole text. First, we shifted and scaled each tilt of~the~noiseless input projections separately according to~the~average mean and standard deviation of~each tilt within the~training set. Through this tilt-dependent scaling, we attempt to~model different degrees of~attenuation as a~function of~tilt angle - the~more extreme the~tilt angle, the~greater the~attenuation due to~the~longer electron beam trajectory.

Next, we simply add Gaussian noise, as this is the~natural choice of~practitioners when they need to~quickly create some simulated tilt-series. Of course, this raises the~question of~how much noise to~add. 
In reality, practitioners would probably visually choose a~value for $\sigma$ that gives a~similar signal-to-noise ratio. However, we opted for a~more objective procedure by subtracting the~noiseless content from the~target projections to~extract the~noise and measure its statistics. With SHREC data, where precise ground truth is available, this task is less challenging than with real data, where we lack ground truth for the~entire tomogram. So in~a~real scenario we would have to~select regions for which we can estimate the~ground truth and calculate the~noise statistics only for these regions instead of~using the~whole tilt values. After adding the~Gaussian noise, the~resulting volume was scaled to~match the~average mean and standard deviation of~the~training set with respect to~tilt. The~steps for creating projections and consequently reconstructions using this method are shown in~\Cref{fig:diagram}, and an example of~a~simulated projection can be seen in~\Cref{fig:projbaseline}. The~calculation of~this modality takes only a~few seconds using a~single CPU.

\ifarxiv
\begin{figure}[!t]
\vskip 0.2in
\begin{center}
\centerline{\includegraphics[width=\columnwidth]{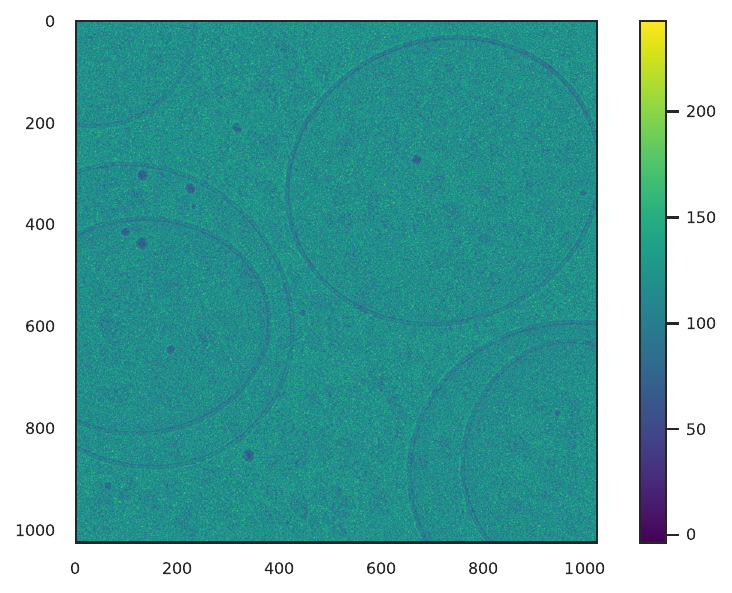}}
\caption{\figProjfaketTitle~\figProjfaketCaption}
\label{fig:projfaket}
\end{center}
\vskip -0.2in
\end{figure}
\fi

\ifarxiv
\begin{figure*}[!t]
\vskip 0.2in
\begin{center}
\centerline{\includegraphics[width=0.98\textwidth]{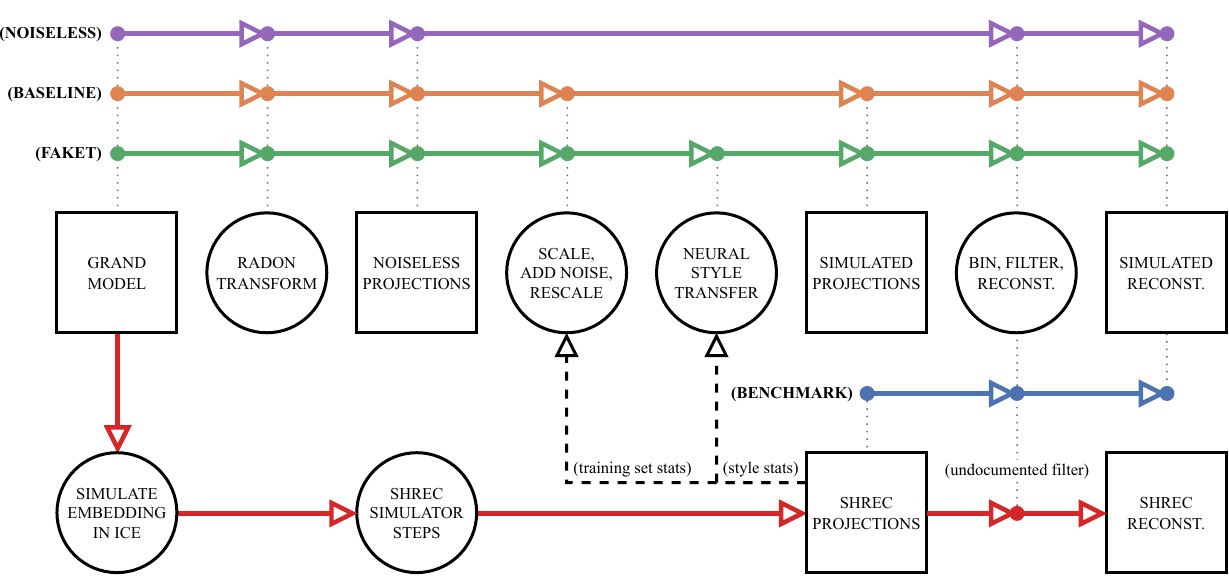}}
\caption{\figDiagramTitle~\figDiagramCaption}
\label{fig:diagram}
\end{center}
\vskip -0.2in
\end{figure*}
\fi

\ifcell \subsubsection*{Neural Style Transfer ({\sc faket})} \else
\subsection{Neural Style Transfer ({\sc faket})} \fi
\label{subsec:nst}
To capture the~noise structure of~the~{\sc benchmark} projections more closely, in~{\sc faket} we devised a~more elaborate method of~estimating the~noise statistics as opposed to~the~simpler one used in~{\sc baseline}. In this case, we estimate the~noise statistics for each tilt in~the~whole training set separately and fit a~second-degree polynomial over the~averages. This process captures the~average noise statistics as a~function of~tilt angle based on~the~information from the~whole training set. After adding the~better estimated Gaussian noise, we obtained projections which we will refer to~as \emph{noisy projections}. To further adapt those projections, we used the~Neural Style Transfer technique implemented in~PyTorch framework\cite{pytorch2019} and introduced in\cite{Gatys_2016_CVPR}. 

NST was built to~render the~semantic content of~natural images in~different styles. At its core lies VGG~net\cite{VGG}, a~convolutional neural network optimized for object recognition and -localization. Within the~NST framework, the~VGG\hl{-19 model} is used for extracting the~content and style representations of~so-called ``content'' and ``style'' images, which are provided as inputs. NST then iteratively updates the~output image to~simultaneously match the~content representation and the~style representation of~the~provided inputs at multiple scales. \hl{The NST technique} is described in~the~aforementioned paper in~great detail and we encourage the~reader to~consult it if any questions would arise. 
\hl{The~VGG-19 model used in {\sc faket} was pretrained on ImageNet data set of natural images, eliminating the need for users to train the model themselves. For those interested in further fine-tuning the NST model (benefits of which are still under investigation by the authors), guidelines are available on the PyTorch website. Fine-tuning would necessitate partially labeled data, unless an autoencoder approach is adopted, which would require no labels.}

Cryo-electron projections are not natural images as those used to~pre-train the~VGG net. It would be therefore surprising if the~NST provided us with desired results ``out-of-the-box''. The~first experiments with NST using the~noiseless projections as content images and {\sc benchmark} projections as style images were disappointing due to~numerous strong artifacts scattered apparently randomly over the~adapted projections. Search over the~space of~hyper-parameters did not result in~satisfactory output even after 10 thousand iterations. Further experiments with {\sc baseline} projections as content images performed poorly on~the~evaluation task, especially on~localization of~smaller particles. It might be useful for the~readers to~know that before we even decided to~work in~projection space, our very first experiments were done in~reconstruction space, but this idea had to~be quickly abandoned due to~the~poor performance, failure to~transfer the~artifacts of~reconstruction algorithm and computational infeasibility.

In order to~obtain the~desired results, we adjusted the~NST to~our specific needs. Firstly, we adjusted the~code to~accept sequences of~1-channel floating point arrays as inputs to~improve the~speed of~processing. Next, we rid the~code of~conversions associated with handling RGB images in~order to~preserve the~floating point precision of~our data. And finally, we implemented the~support for other than random initialization or initialization with the~content which was a~crucial change to~produce the~desired results. We initialized the~NST with \emph{noisy projections}, used \emph{noisy projections} with 25\,\% of~noise as content, and provided the~associated {\sc benchmark} projections as style images. We would like to~point out that the~associated style images were taken from a~different training tomogram, therefore do not feature the~same content. I.e.~it is not possible to~simply minimize the~element-wise mean absolute error to~get the~desired output.  

From the~NST pipeline, we only used the~1024x1024 scale because downsampling to~smaller scales combined with anti-aliasing used in~NST to~transition between the~scales is well suited for natural images and lots of~NST iterations but not for our use case on~scientific data. Using new initialization and slightly noisy content images allowed us to~increase the~learning rate of~the~NST optimizer, so we were able to~produce visually persuasive outputs in~just one iteration, as seen in~\Cref{fig:projfaket}.

This method is more involved than just adding the~noise, but produces better results, is very fast due to~our changes of~NST initialization and hyper-parameters, requires less domain-specific knowledge than implementing the~forward operator, and can be used in~real-world scenarios. Computing this modality of~data, including the~steps to~create the~\emph{noisy projections} and \emph{content projections} for 10 tomograms took only $\approx$12 minutes on~a~single \emph{NVIDIA A100 40GB SXM4} GPU, which on~average represents a~$750\times$ speedup in~comparison to~{\sc benchmark}. This method also needed only $\approx$3.5\,GB of~GPU memory per tomogram as opposed to~117GB of~RAM which was necessary for {\sc benchmark}. The~GPU memory usage of~our method could be even further optimized. This represents 33 times less memory usage. Certainly, it is not possible to~do a~head-to-head comparisons between GPU memory and RAM, but this information can be used to~estimate the~hardware requirements. Moreover, as opposed to~the~{\sc benchmark} method, it is possible to~scale our method to~volumes that match usual sizes of~experimental data on~still reasonable hardware. 

\else
\refstepcounter{figure}\label{fig:projbaseline}
\refstepcounter{figure}\label{fig:projfaket}
\refstepcounter{figure}\label{fig:diagram}
\fi

\ifarxiv
\section{Evaluation}
\label{sec:evaluation}
To evaluate the~quality of~the~proposed simulated projections and subsequently tomograms, no widely accepted metrics exist\cite{seemann2020data}, and it is not clear how well the~standard metrics such as the~mean squared error, or more elaborate metrics such as the~Fr\'{e}chet inception distance\cite{NIPS2017_FIDref}, relate to~the~performance on~tasks of~interest to~the~practitioners.

We therefore trained a~randomly initialized DF neural network to~solve two evaluation tasks proposed in~the~SHREC 2021 challenge, namely particle localization and classification. These tasks are of~utmost importance to~practitioners in~determining the~structure of~proteins and macromolecular complexes. We argue that evaluation on~a~task of~practical relevance is much more valuable than merely computing the~currently available metrics. It also gives the~reader an advantage to~see the~newly created data modalities in~practice, even though the~computational cost of~evaluation is high. \ifcell Nevertheless, we also provide the evaluation results using the standard metrics in \Cref{table:standardmetrics}.\fi

To obtain the~main results, we trained DF for 70 epochs on~{\sc benchmark}, {\sc faket}, and {\sc baseline} training data, every time across 6 different seeds of~randomness ($3\times 6$ times in~total). Each model was tested on~the~same test tomogram -- \emph{model\_9} from the~original SHREC data set. The~final comparison was done by observing the~models' performances in~terms of~F1 score for localization task and F1 macro score for classification task. The~computation of~the~scores was done the~same way as in\cite{shrec2021} where also further particularities of~the~tasks are described. To estimate the~68\,\% and 95\,\% confidence intervals, we used the~bootstrapping method. The~results are visualized and discussed in
\ifarxiv \Cref{sec:results}. \fi
\ifcell \nameref{sec:results}. \fi
Results of~additional experiments are presented in~the~\ifarxiv Appendix\else Supplementary material\fi.

\ifcell \subsubsection*{DeepFinder} \else
\subsection{DeepFinder} \fi
\label{subsec:deepfinder}

DeepFinder emerged from the~SHREC challenge as one of~the~most successful candidates. One other reason why we chose to~use DF was the~availability of~its source code. DF is a~deep 3D-convolutional neural network trained in~a~supervised fashion by optimizing a~dice loss. For training of~the~models, we used the~Adam optimizer with learning rate of~$0.0001$. The~exponential decay rate was set to~$0.9$ for the~first moment estimate and $0.999$ for the~second moment estimate as in\cite{Deepfinder}. 

To train the~DF, we used tomograms containing particle instances from 16 classes (two of~those being background and vesicles). We adjusted the~training procedure of~DF such that in~one epoch, all patches containing particles within the~training data set were seen once, or with minimal repetition. a~patch in~a~shape of~a~box is rotated by 180$^{\circ}$ at random. Patches are processed in~batches constructed from a~randomly permuted list of~all particles in~the~training set. After our changes in~the~training procedure, an epoch represents a~different number of~gradient steps than in\cite{Deepfinder}, therefore the~numbers of~epochs are not comparable with the~original paper. To perform the~particle localization and -classification on~the~test tomogram, DF carries out a~segmentation procedure followed by clustering and thresholding steps, where the~two latter steps are computed using a~CPU. The~thresholding is used to~reduce the~number of~false positive findings.

To train the~DF network implemented in~Keras framework\cite{chollet2015keras} using TensorFlow backend\cite{tensorflow2015}, we utilized multiple \emph{NVIDIA A100 40GB SXM4} GPUs. That allowed us to~run multiple experiments in~parallel. One training job comprising of~9 training tomograms required approx. 17GB of~GPU memory, therefore we were able to~submit two jobs on~one GPU at a~time. One training epoch, as defined earlier, took $\approx$21\,min.\@ to~finish. The~segmentation of~a~test tomogram took $\approx$2.5\,min.\@ on~the~aforementioned GPU. The~clustering step took $\approx$30\,min.\@ on~a~single core of~a~\emph{$2\times64$-core AMD EPYC 7742} CPU. The~most computational expensive step was thus the~training of~the~DF. 
\fi

\section{Results}
\label{sec:results}

Comparison of~DeepFinder’s performance on~localization and classification tasks as a~function of~training data is presented in~\Cref{table:result}. The~mean performances per epoch with 68\,\% and 95\,\% confidence intervals (CI) are presented in~\Cref{fig:localization} and \Cref{fig:classification}. Results of~additional experiments can be found in~\Cref{appendix:finetuning,,appendix:additionalexps,,appendix:confmats}.

\def \figLocalizationTitle {Performance on~localization task.}
\def \figLocalizationCaption {Darker and lighter regions show 68\% and 95\% CI.}
\ifarxiv
\begin{figure}[!t]
\vskip 0.2in
\begin{center}
\centerline{\includegraphics[width=\columnwidth]{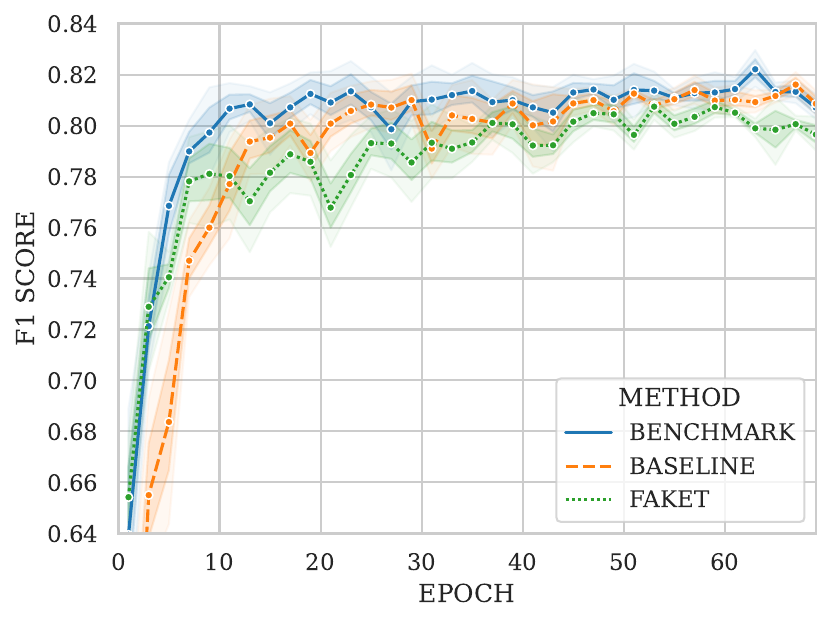}}
\caption{\figLocalizationTitle~\figLocalizationCaption}
\label{fig:localization}
\end{center}
\vskip -0.3in
\end{figure}
\else
\refstepcounter{figure}\label{fig:localization}
\fi

The performance of~DF trained on~{\sc benchmark} data turned out to~be the~best in~both localization and classification tasks, as expected, since the~testing tomogram was reconstructed from the~exactly same tilt-series as {\sc benchmark} training set. In practice, however, it is infeasible to~have such favorable conditions as fully labelled tomograms are not existent. Most biological studies focus on~one or a~subset of~molecules to~be labelled. And since the~labelling process is laborious and requires domain experts, it is also expensive. With that in~mind, we can hardly expect to~have a~training set of~tomograms from exactly the~same distribution as the~tomograms we are interested in. Nevertheless, this result serves us, as the~name suggests, as a~benchmark that we are trying to~reach with our proposed methods. 

The highest {\sc benchmark} F1 score of~81.5\,\% (68\,\%~CI [81.1\,\%, 82.0\,\%]) on~localization task and 58.1\,\% (68\,\%~CI [57.9\,\%, 58.3\,\%]) on~classification task on~average across 6 different random seeds was achieved after 65$\pm$4 epochs of~training. To put these numbers in~perspective, we investigated the~limits of~DF performance by training and evaluating on~completely noiseless data, see~\Cref{appendix:additionalexps} for more details. In this setting DF achieved F1 score of~83.2\,\% (68\,\%~CI [82.4\,\%, 84.0\,\%]) on~localization task and 72.5\,\% (68\,\%~CI [70.5\,\%, 74.6\,\%]) on~classification task.

\def \figClassificationTitle {Performance on~classification task.}
\def \figClassificationCaption {Darker and lighter regions show 68\% and 95\% CI.}
\ifarxiv
\begin{figure}[!t]
\vskip 0.2in
\begin{center}
\centerline{\includegraphics[width=\columnwidth]{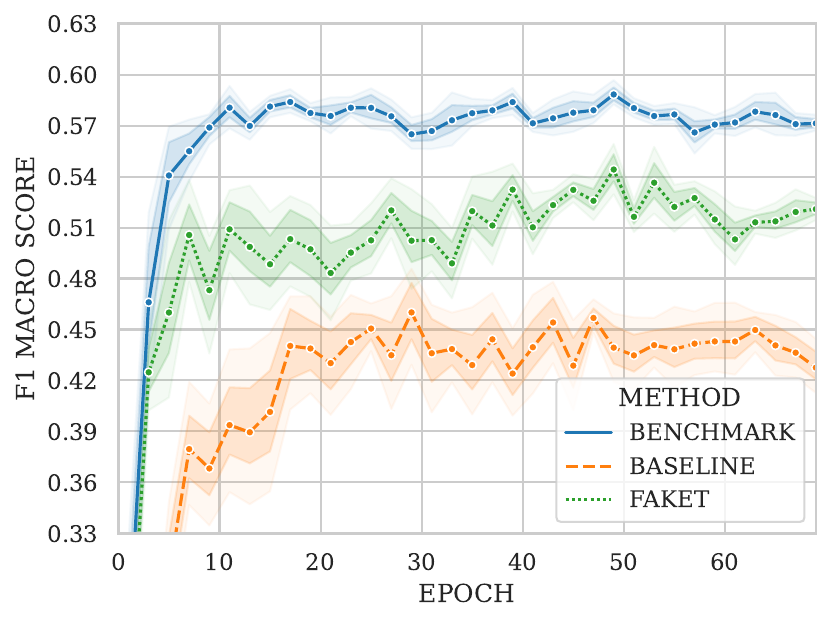}}
\caption{\figClassificationTitle~\figClassificationCaption}
\label{fig:classification}
\end{center}
\vskip -0.3in
\end{figure}
\fi 

Our fastest method of~simulating the~projections and subsequently the~tomograms, {\sc baseline}, scored 81.3\,\% (68\,\%~CI [81.1\,\%, 81.6\,\%]) on~localization task. It was a~surprise that such a~simple method led to~so high localization performance, considering that it requires $\approx 150\,h$ to~create the~{\sc benchmark} projections, while it takes almost no time to~create the~{\sc baseline} projections. This method can be therefore regarded as the~``poor-man's'' choice in~settings where the~computational budget is very limited, or in~settings where this task has to~be performed repeatedly many times. Unfortunately, the~classification performance of~44.1\,\% (68\,\%~CI [43.2\,\%, 45.1\,\%]) was rather poor, not surprisingly, as classification is an inherently harder task, cf.\cite{shrec2021}\ifarxiv.\fi

\def \figPerclassTitle {Per-class performance on~classification task.}
\def \figPerclassCaption {Darker and lighter regions show 68\% and 95\% CI.}
\ifarxiv
\begin{figure*}[!t]
\vskip 0.2in
\begin{center}
\centerline{\includegraphics[width=\linewidth]{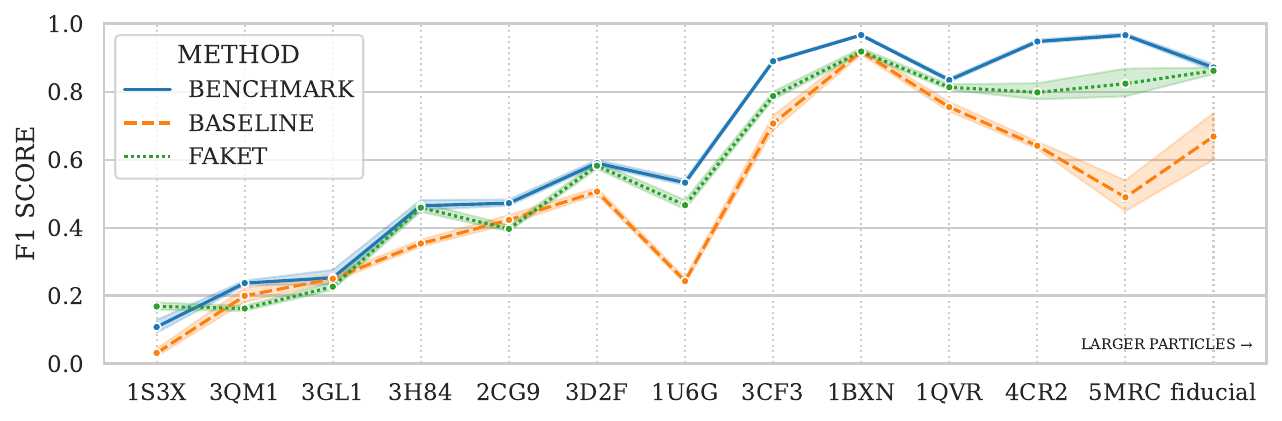}}
\caption{\figPerclassTitle~\figPerclassCaption}
\label{fig:perclass}
\end{center}
\vskip -0.2in
\end{figure*}
\fi

Our proposed method {\sc faket}, based on~additive noise (using our advanced estimation of~in-projections noise) and subsequently neural style transfer, with on~average 80.0\,\% (68\,\%~CI [79.8\,\%, 80.2\,\%]) F1 score on~localization task performed on~par with the~{\sc benchmark} and significantly outperformed standard template matching algorithms. The~best performance on~the~classification task was achieved after 65$\pm$4 epochs with a~score of~53.3\,\% (68\,\%~CI [52.8\,\%, 53.8\,\%]). With this result, {\sc faket} matched the~{\sc benchmark} to~92\,\% while reducing the~cost of~data generation by a~factor of~750 and using 33 times less memory. Multi-class classification performance certainly can not be reduced only to~one number and it is important to~consider also other metrics, such as per-class classification report and confusion matrices. The~per-class classification performance is shown in~\Cref{fig:perclass} and we provide the~confusion matrices in~\Cref{appendix:confmats}. The~full performance reports are available in~the~accompanying repository for enthusiastic readers. 

In~\Cref{appendix:finetuning}, we present further experimental results showcasing the~fine-tuning of~{\sc faket} models using a~subset of~{\sc benchmark} data. This approach further narrows the~performance gap between the~{\sc faket} and {\sc benchmark} models, achieving 97\,\% of~the~{\sc benchmark} model's classification performance, while exceeding its performance on~localization task. These results might~be valuable for cryoET practitioners who are seeking to~maximize the~performance of~their models and have, or can get a~small but representative sample of~their data labeled.



\ifarxiv\balance\fi
\section{Discussion}
\label{sec:discussion}

In this paper, we proposed {\sc faket}, a~\hl{fast and scalable data-driven} method for simulating the~forward operator of~an\hl{y} electron microscope based on~additive noise and Neural Style Transfer. The~proposed method can be used for generating synthetic cryo-electron \hl{micrographs or tilt-series that closely approximate the quality of reference TEM data, at a~computational cost only marginally higher than that of simply adding Gaussian noise. The data} can be used to~train deep neural networks to~solve tasks such as particle localization and (much more challenging) particle classification. The~field of~cryo-electron tomography currently suffers from the~lack of~sufficient amounts of~annotated data, and the~proposed method aims to~solve this problem. 

\hl{In this study, we evaluated} {\sc faket} as a~surrogate model that mimics the~behavior of~the~physics-based TEM simulator SHREC, while being drastically less computationally expensive, both in~terms of~time and memory. It accelerates the~data generation process by a~factor of~750 while using 33 times less memory, making the~generation of~thousands of~tilt-series feasible. Moreover, it is capable of~simulating large tilt-series, which are common in~experimental environments. For example, we generated a~$61\times3500\times3500$ tilt-series on~a~single \emph{NVIDIA A100 40GB SXM4} GPU in~less than 10 minutes. It also does not require any calibration protocol, it does not need to~be retrained to~be used on~new data, nor does it require labeled reference data. The~method is open source, and the~experiments are reproducible. 

The quality of~our approximations was evaluated using the~DeepFinder network, which emerged from the~SHREC challenge as one of~the~most successful. The~results showed that the~performance of~models trained using our approximations is on~par with the~{\sc benchmark} method on~localization task and reached 92\,\% of~its performance on~classification task while significantly outperforming standard template matching algorithms. When further fine-tuned using a~portion of~{\sc benchmark} data, the~classification performance was improved to~97\,\%. 

This advancement simplifies the~generation of~fully-labeled, high-quality synthetic tilt-series that resemble experimental TEM data requiring analysis. This simulated data can either be used to~train particle localization and classification neural networks from scratch, or serve as pre-training data for networks that will be fine-tuned with manually labeled experimental data later. Researchers investigating reconstruction algorithms can also benefit from our simulator since the~availability of~ground truth allows for effortless evaluation of~their novel methods. \hlb{However, when simulating data for use cases other than training neural networks for particle localization or classification, it is important to take into account that no explicit knowledge of the underlying physics phenomena of TEM is built into the simulator.}

In future work, we will focus on~further validation of~our method using experimental \hlb{TEM data.} Additionally, we plan to~improve the~method by \hlb{replacing the VGG-19 network that was pretrained on natural images, with a state-of-the-art vision network pretrained on cryoEM data. Moreover, we aim} to~provide the~community of~practitioners with a~fully functional and easy-to-use piece of~software \hlb{for~generating their synthetic data sets. Either} based on~chosen particles from Protein Data Bank\hlb{, or using already available whole-cell models}. The~goal is to~enable more accurate and efficient data analysis while also making the~process more accessible to~researchers in~the~field. \hlb{We hope this advancement will serve as the~basis for development of new computational methods in~cryoEM and cryoET.}

\clearpage

\balance
\ifarxiv
\bibliography{main}
\fi

\ifarxiv
\newpage
\appendix
\onecolumn
\clearpage
\ifarxiv
\section{Side-by-side comparison of~projection types}
\else
\section{Comparison of~projection types related to STAR Methods}
\fi
\label{appendix:sidebyside}

\begin{figure}[h]
\begin{center}
\centerline{\includegraphics[width=0.79\textwidth]{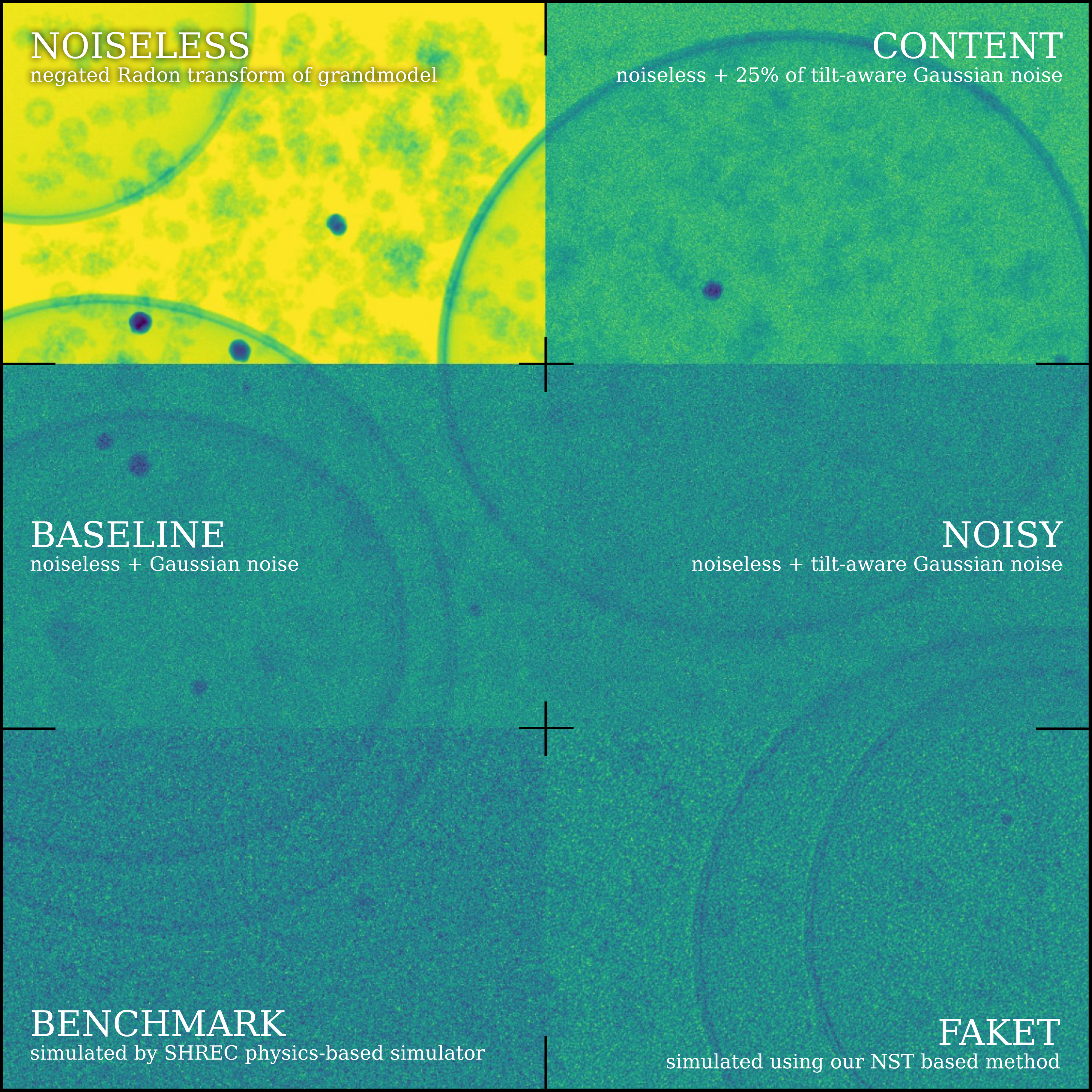}}
\caption{Side-by-side comparison of~all types of~projections presented in~this article\ifcell , related to STAR Methods\fi.}
\label{fig:side-by-side}
\end{center}
\end{figure}
\begin{figure}[H]
    \vspace{-14mm} 
    \centering
    \subfloat{
        \includegraphics[height=0.21\paperheight]{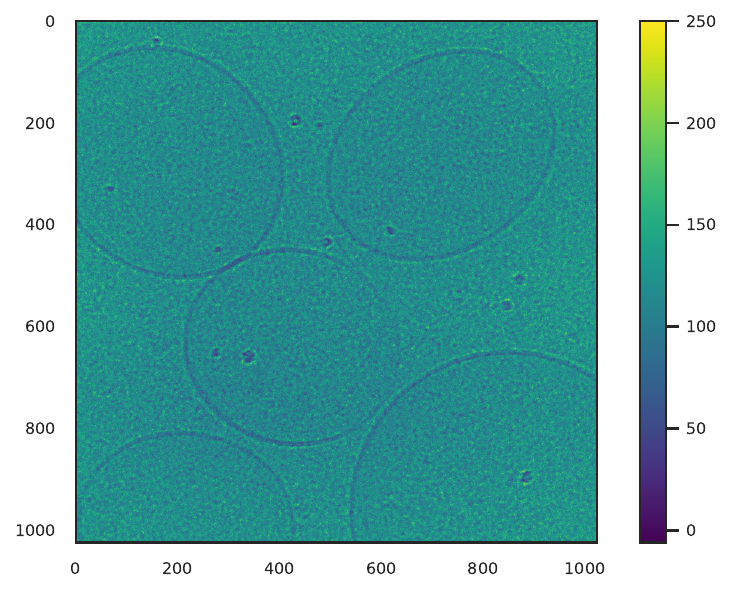} }
    \subfloat{
        \includegraphics[height=0.21\paperheight]{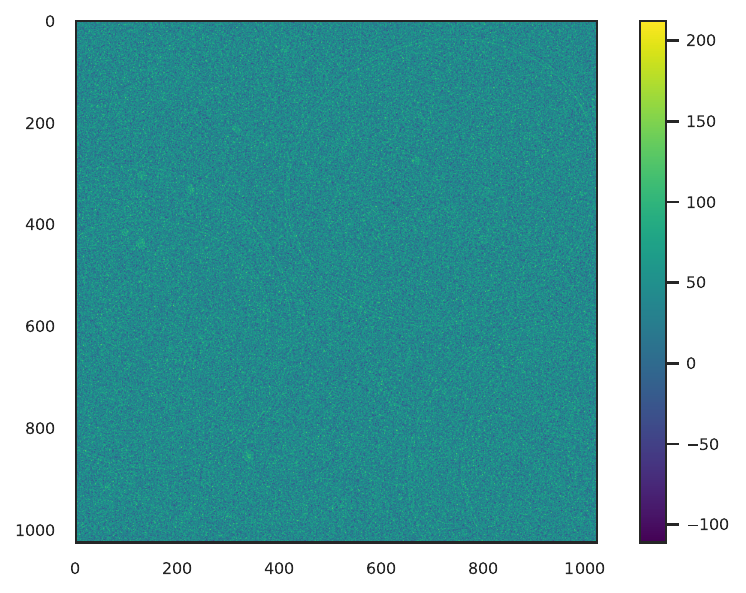} }
    \caption{One instance of~style projections (from {\sc benchmark})  used as a~reference for NST (left) and the~difference between one particular instance of~{\sc faket} and {\sc benchmark} projections\ifcell , related to \Cref{fig:projfaket}\fi. Note that the~style projection used as a~reference input to~NST never features the~same content as the~input projection that is being style-transferred. \hlb{Additional comparisons of projections \ifarxiv through visualizations and standard metrics are presented in the supplement\else are presented in \Cref{table:standardmetrics,supfig:inputs1,supfig:inputs2,supfig:inputs3}\fi. Colorbar indicates simulated intensities in arbitrary units.}}%
    \label{fig:teaser}%
    \vspace{-5mm}
\end{figure}


\clearpage
\ifarxiv
\section{Recreating {\sc benchmark} -- reverse-engineering SHREC filtering step}
\else
\section{Recreating {\sc benchmark} -- reverse-engineering SHREC filtering step related to \Cref{fig:diagram}}
\fi
\label{appendix:benchmark}

In pursuit of~re-creating the~{\sc benchmark} reconstructions to~match those from SHREC data set as closely as possible, we inspected the~public source code of~PyTom\cite{pytom} for clues on~which filters might have been used. In the~original paper, authors only mention that they use weighted back-projection. Which would suggest the~use of~a~simple ramp filter. In a~personal communication, they mentioned using a~default filter from their private PyTom version with undefined Crowther frequency. Therefore, we implemented and used a~simple ramp filter with a~support for Crowther frequency. Then, we visualized SHREC’s and our reconstructions in~Fourier space to~visually compare the~results. Unfortunately, none of~the~variations of~a~simple ramp filter, even with Crowther frequency, produced the~expected result similar to~SHREC.  
For that reason, we started to~iteratively change our filter in~order to~approximately match the~aforementioned reconstructions in~both real and Fourier space, in~hope of~finding the~desired match, see~\Cref{fig:middleZ}. After many iterations, we found a~good-enough match, a~filter which is a~product of~a~2D Gaussian filter ($\sigma_{x} = 174$, $\sigma_{y} = 102$), a~1D ramp filter broadcasted to~2D ($\mathrm{CrowtherFreq} = 0.61$~Nyquist), and a~2D circular filter ($\mathrm{radiusCutoff} = 256$, i.e.~Nyquist frequency). Exact implementation is available in~code repository.

\begin{figure}[h]
    \centering
    \subfloat[\centering Reconstruction space]{{\label{fig:middleZreal}
        \includegraphics[width=0.98\textwidth]{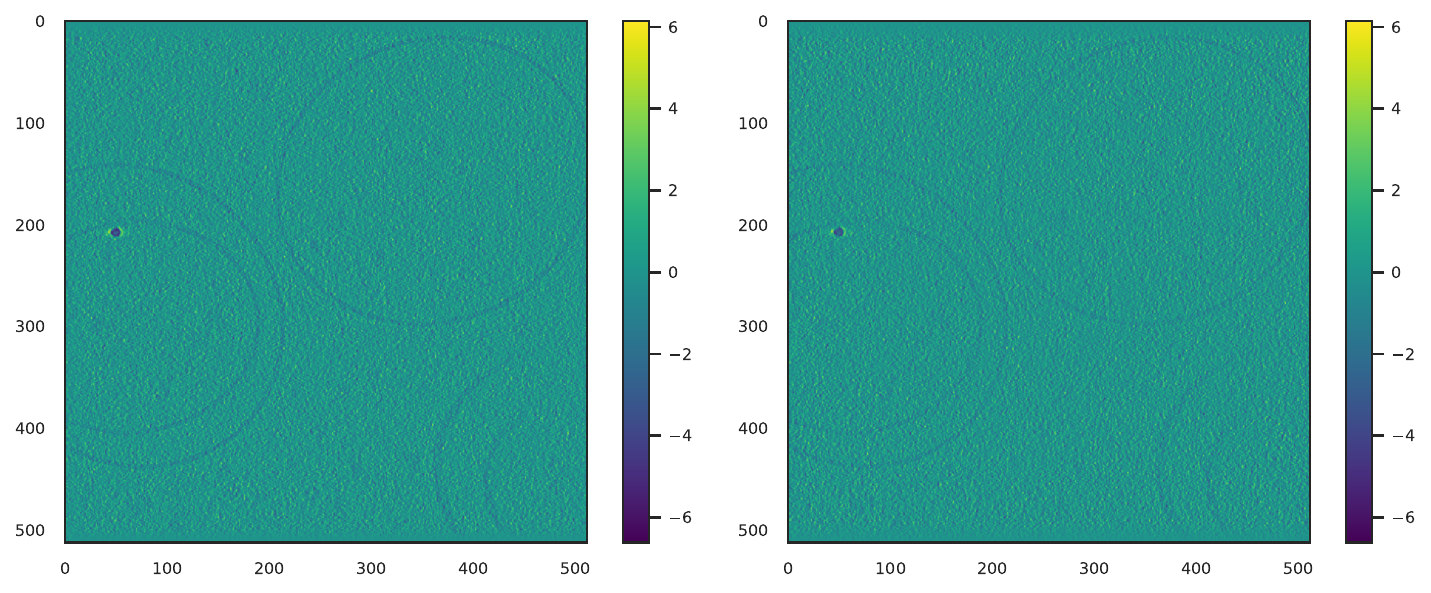} }}%
        \hspace{3mm}
    \subfloat[\centering Fourier space]{{\label{fig:middleZfourier}
        \includegraphics[width=0.98\textwidth]{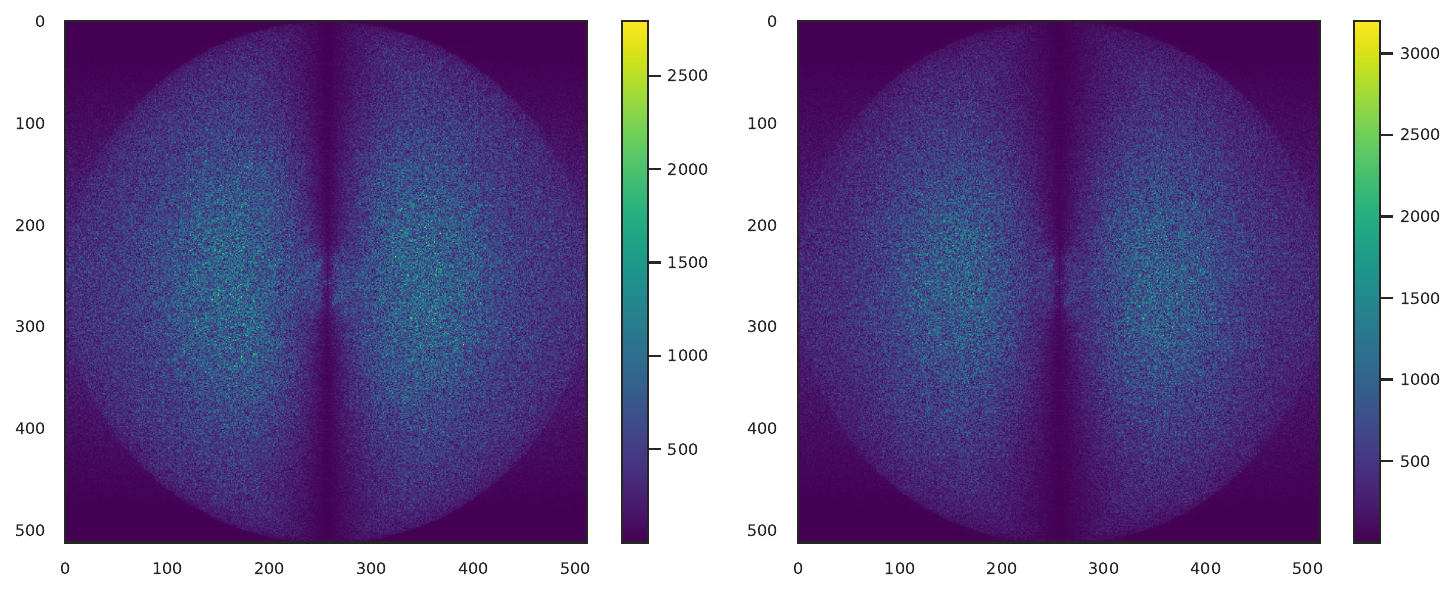} }}%
    \caption{Middle Z-axis slice of~SHREC (left) and {\sc benchmark} (right) reconstructions\ifcell , related to STAR Methods\fi. Colorbar indicates simulated intensities in arbitrary units (top), and absolute value of the Fourier transform (bottom).}%
    \label{fig:middleZ}%
\end{figure}


\clearpage
\ifarxiv
\section{Further improving the~performance of~DF models trained with {\sc faket} data by subsequent fine-tuning on~a~fraction of~{\sc benchmark} data}
\else 
\section{Further improving the~performance of~DF models trained with {\sc faket} data by subsequent fine-tuning on~a~fraction of~{\sc benchmark} data, related to STAR Methods}
\fi
\label{appendix:finetuning}

Fine-tuning is a~prevalent approach in~machine learning where a~pre-trained model is further refined using a~subset of~the~available data specific for the~task. This methodology often yields improved performance by allowing the~model to~specialize on~the~nuances of~a~specific data set. For practitioners in~cryoET, this process can be of~great value when seeking to~maximize the~performance of~their models. If they are able to~label a~small but representative sample of~their data manually, they can leverage this technique to~their advantage. 

Here, we compared the~performance of~models trained on~{\sc baseline} or {\sc faket} data with that of~a~model initially trained on~{\sc faket} data and later fine-tuned using a~subset (3 out of~9) of~{\sc benchmark} tomograms, denoted as {\sc finetuned-3} in~the~\Cref{fig:finetune}. The~fine-tuning began after the~50$^{th}$ epoch, triggering a~notable improvement in~both localization and classification performance. To make a~reasonable comparison, we include results from a~model, {\sc benchmark-3}, trained solely on~the~same 3 {\sc benchmark} tomograms used for fine-tuning.

During the~fine-tuning and {\sc benchmark-3} training stages, the~total number of~training tomograms was smaller. To maintain experiment consistency by keeping an approximately equivalent number of~gradient steps per epoch, we adapted our definition of~an epoch. Here, it refers to~a~cycle where the~model is trained on~9 tomograms — meaning that during fine-tuning, three full training cycles with 3 tomograms each are considered one epoch.

\begin{figure}[h]
    \centering
    \subfloat[\centering Performance on~localization task]{{\label{fig:finetune-a}
        \includegraphics[height=0.193\paperheight]{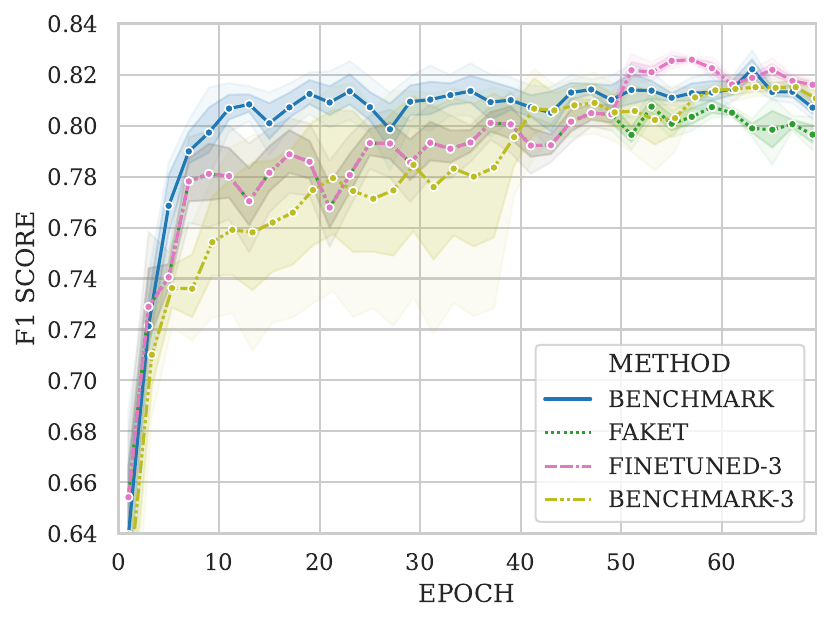} }}%
        \hspace{3mm}
    \subfloat[\centering Performance on~classification task]{{\label{fig:finetune-b}
        \includegraphics[height=0.193\paperheight]{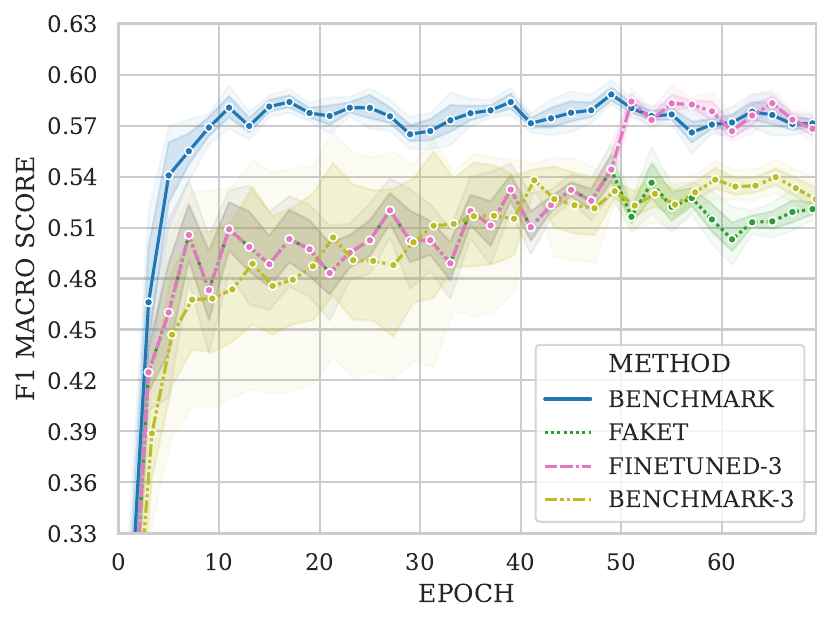} }}%
    \\
    \hspace{-2mm}\subfloat[\centering Per-class performance on~classification task]{{\label{fig:finetune-c} 
    \includegraphics[width=0.915\linewidth]{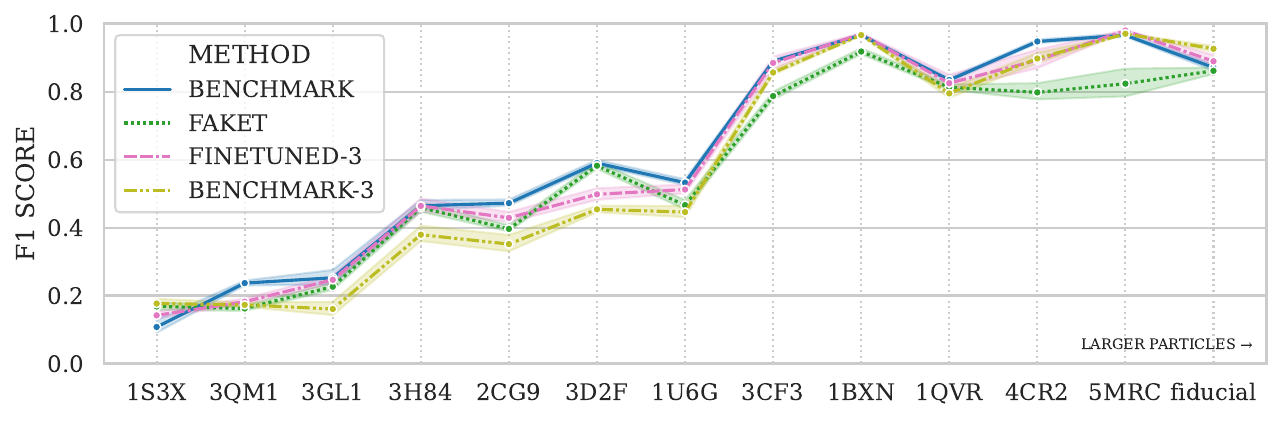} }}%
    \caption{Performance of~{\sc finetuned-3} and {\sc benchmark-3} models in~context with the~main results\ifcell , related to STAR Methods\fi. Darker and lighter regions show 68\% and 95\% CI.}
    \label{fig:finetune}%
\end{figure}


\clearpage
\ifarxiv
\section{Additional experiments}
\else
\section{Additional experiments, related to STAR Methods}
\fi
\label{appendix:additionalexps}

For our {\sc baseline}, we chose the~conventional approach of~adding Gaussian noise. However, acknowledging the~noise component in~TEM projections fluctuates with tilt angle, we devised a~tilt-dependent algorithm for Gaussian noise addition. This formed the~basis for our NST-based {\sc faket} method, producing what we refer to~as \emph{noisy projections}. One might wonder if such simulation enhances DF performance, a~query best illustrated by a~figure. Although localization performance remained consistent, we observed a~slight enhancement in~classification, as shown in~\Cref{fig:additional} (top). Despite this modest improvement, we didn't deem it significant enough to~modify our simple {\sc baseline} method, but still opted to~incorporate it into our {\sc faket} method.

We present an ablation study featuring a~\emph{faket-random} model trained on~data created with a~randomly initialized NST network, in~contrast to~pre-trained. Results in~\Cref{fig:additional} (middle) underscore the~necessity of~pre-training for successful simulation and suggest that pre-training using cryoET data might further enhance the~simulated data's resemblance to~real tomograms — a~question still under exploration.

Finally, to~offer insight into the~potential limits of~DF on~the~studied data, we share results from a~\emph{noiseless} experiment in~\Cref{fig:additional} (bottom). The~model was trained and tested on~completely noiseless reconstructions, making it also the~only model in~this paper evaluated on~a~noiseless test tomogram.

\begin{figure}[H]
    \centering
    \vspace{-5mm}
    \subfloat{\includegraphics[height=0.193\paperheight]{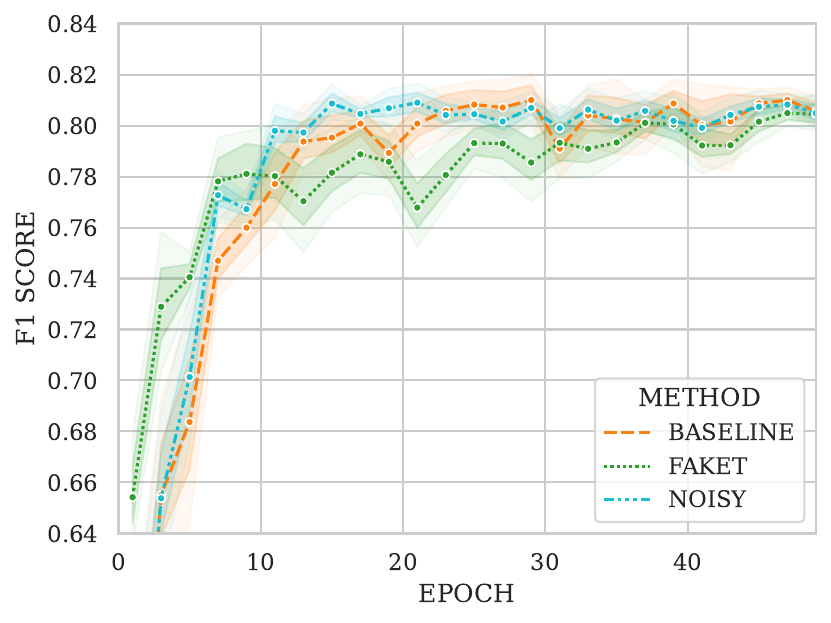} }%
        \hspace{3mm}
    \subfloat{\includegraphics[height=0.193\paperheight]{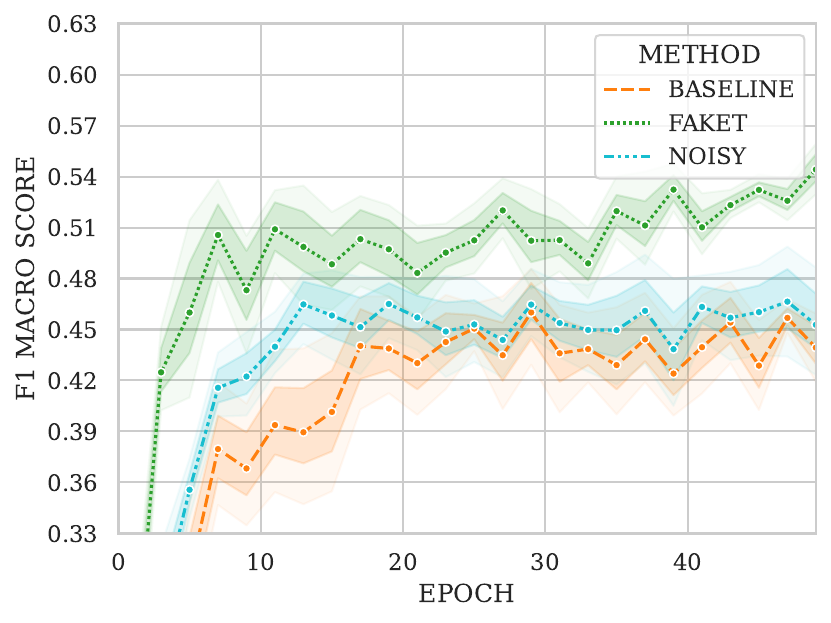} }%
    \\ \vspace{-8.5mm}
    \subfloat{\includegraphics[height=0.193\paperheight]{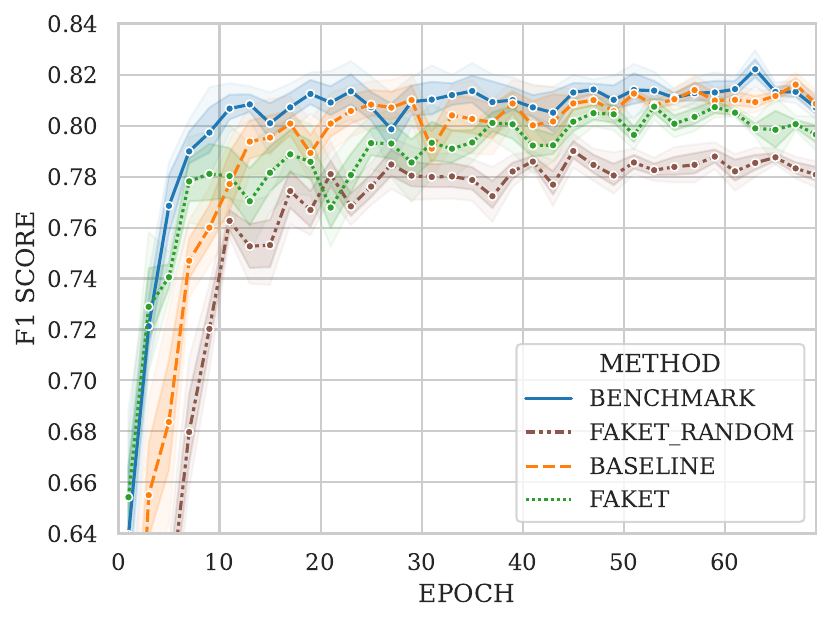} }%
        \hspace{3mm}
    \subfloat{\includegraphics[height=0.193\paperheight]{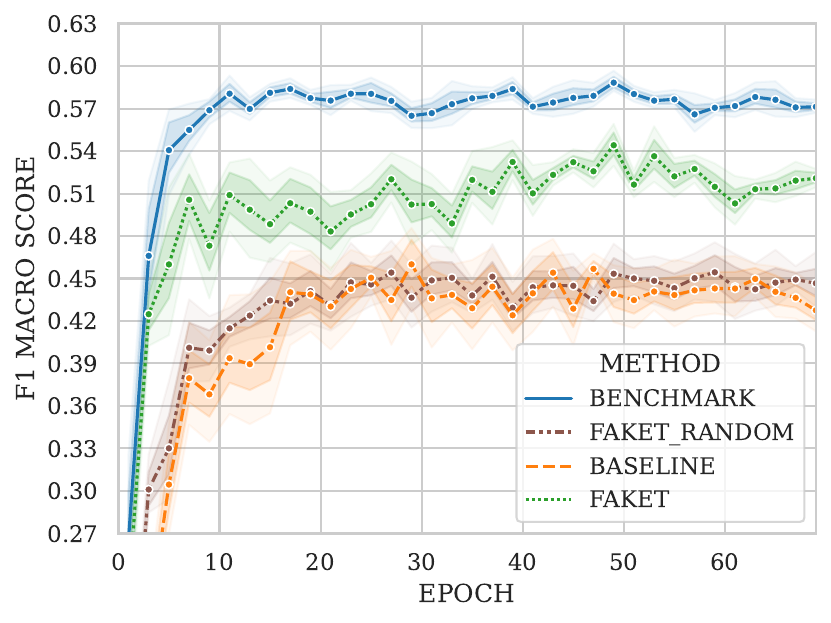} }%
    \\ \vspace{-8.5mm}
    \subfloat{\includegraphics[height=0.193\paperheight]{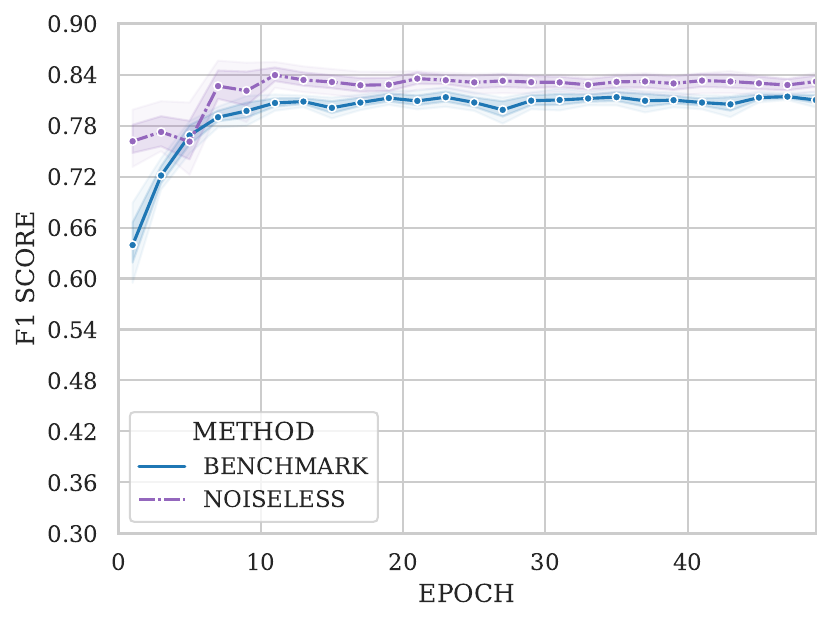} }%
        \hspace{3mm}
    \subfloat{\includegraphics[height=0.193\paperheight]{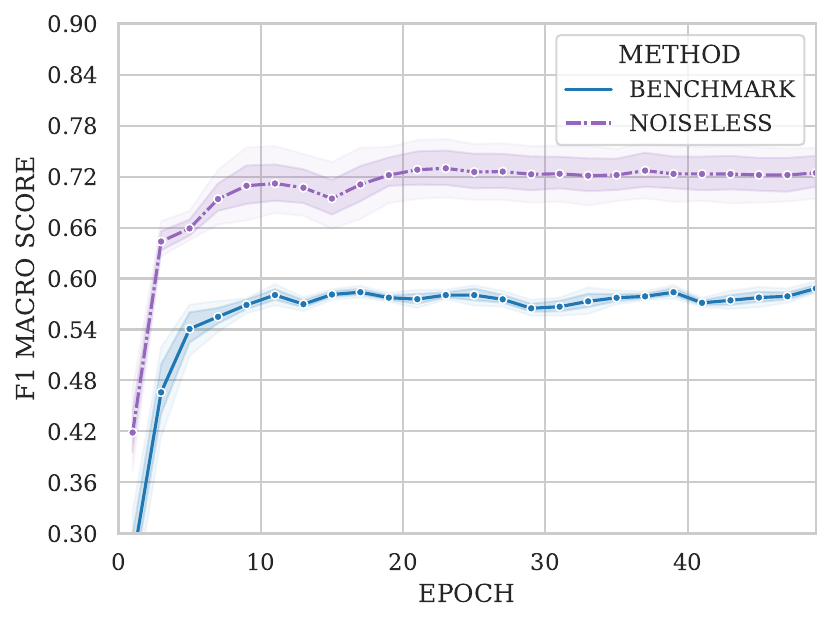} }%
    \caption{Performance of~DF trained with \emph{noisy} (top), \emph{faket-random} (middle), and \emph{noiseless} (bottom) data\ifcell , related to STAR Methods\fi. Results of localization (left) and classification (right). Darker and lighter regions show 68\% and 95\% CI.}
    \label{fig:additional}%
\end{figure}


\clearpage
\ifarxiv
\section{Further details on~particle classification performance}
\else
\section{Detailed classification performance, related to STAR Methods}
\fi
\label{appendix:confmats}

Next to~F1 scores reported in
\ifarxiv \Cref{sec:results}, \fi
\ifcell \nameref{sec:results}, \fi
here we present aggregated confusion matrices for all main particle classification and selected additional experiments\hlb{, see \Cref{fig:confmats_benchmark_faket,fig:confmats_baseline_noiseless,fig:confmats_finetuned-3_benchmark-3}}. Each confusion matrix represents the~average of~6 confusion matrices obtained from 6 different experiments each initialized with a~different random seed. Each confusion matrix was computed based on~the~results of~the~best epoch that was selected based on~the~model's validation performance. The~csv files containing information about which epochs were selected for each of~the~aggregated confusion matrices are available in~the~accompanying repository.  

Particles on~X and Y axes are sorted according to~their volume from the~smallest on~the~left (or top) to~the~largest on~the~right (or bottom). Particles are also grouped into size-groups so the~reader can get a~better feeling for the~relationship between particle volume and DF performance. The~values at the~diagonal represent the~fraction of~correctly classified particles for a~class at any given row. The~values in~parentheses in~each cell show $10^{th}$ and $90^{th}$ percentiles, since the~displayed values are averages. The~values that are not at the~diagonal represent the~fraction of~the~particles that were misclassified as a~member of~the~class at the~particular column. The~sum of~all values in~one row is therefore equal to~100\,\%. Tick labels of~the~Y axis additionally contain information about the~total number of~instances of~each class within the~test tomogram.

\begin{figure}[h]
    \centering
    \subfloat{{\label{fig:confusion_matrix_benchmark}
        \includegraphics[width=0.98\linewidth]{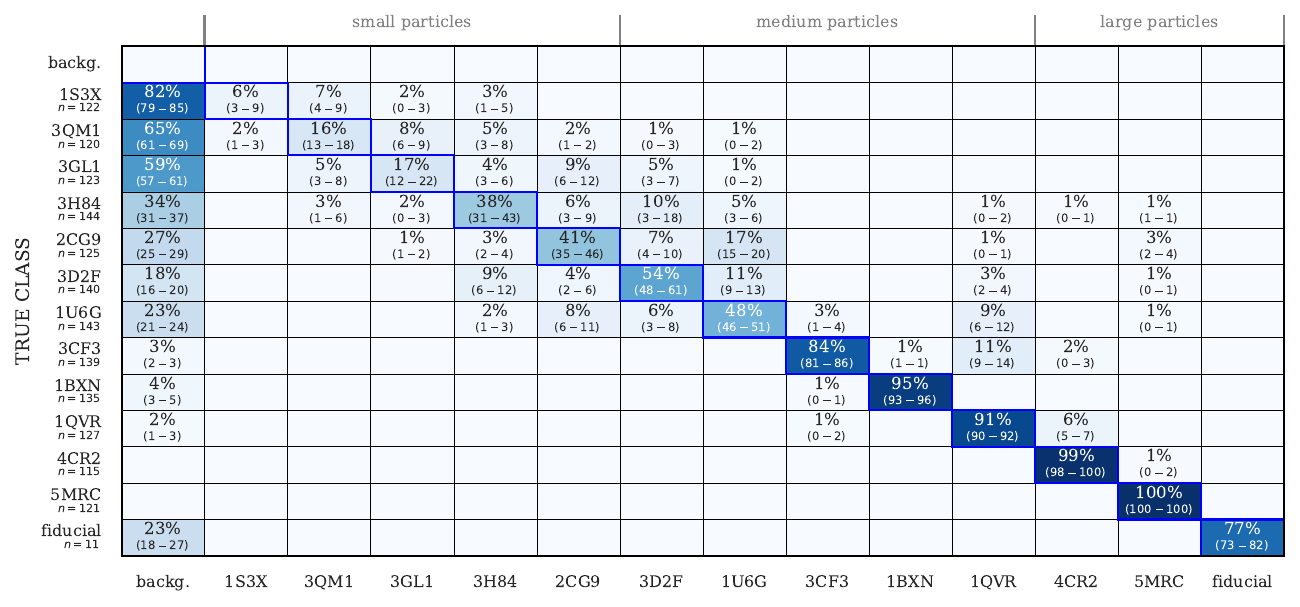} }}%
            \\
    \subfloat{{\label{fig:confusion_matrix_faket}
        \includegraphics[width=0.98\linewidth]{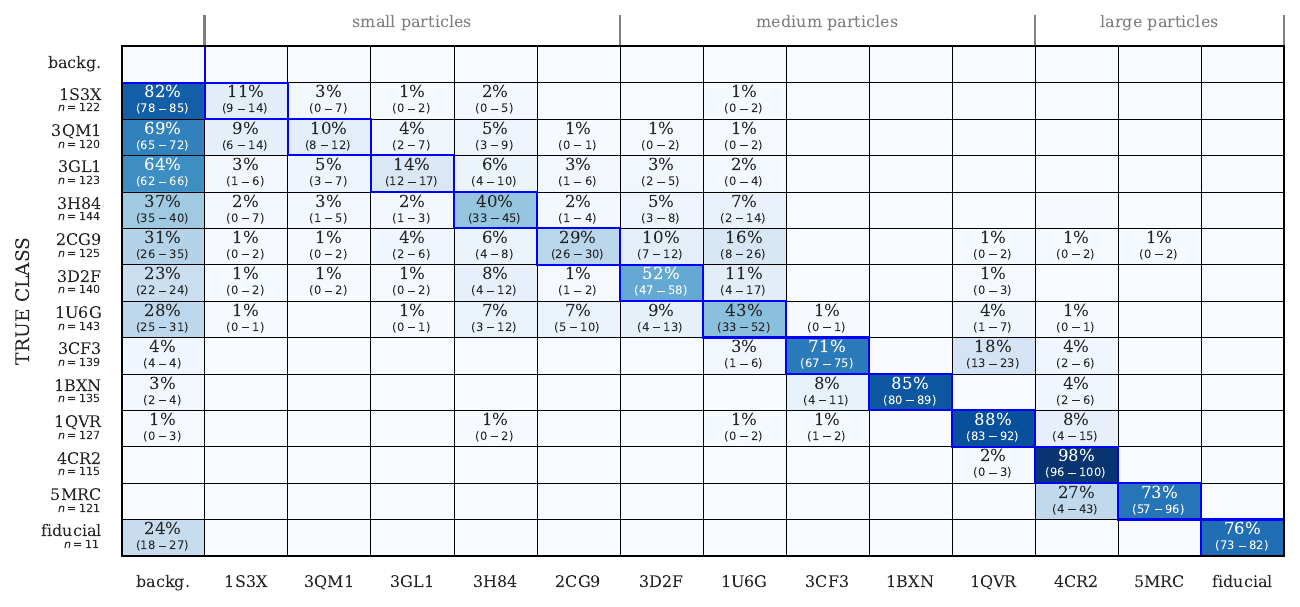} }}%
    \caption{Confusion matrices for {\sc benchmark} (top) and {\sc faket} (bottom) experiments\ifcell , related to STAR Methods\fi.}
    \label{fig:confmats_benchmark_faket}%
\end{figure}

\begin{figure}
    \centering
    \subfloat{{\label{fig:confusion_matrix_baseline}
        \includegraphics[width=0.98\linewidth]{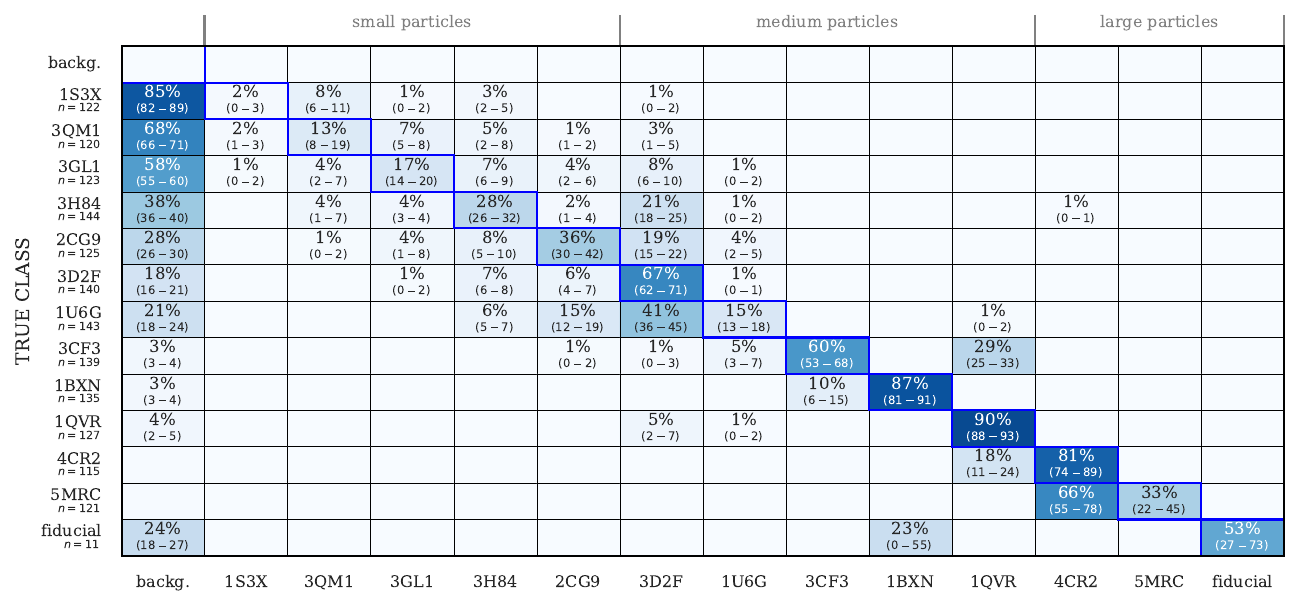} }}%
            \\
    \subfloat{{\label{fig:confusion_matrix_noiseless}
        \includegraphics[width=0.98\linewidth]{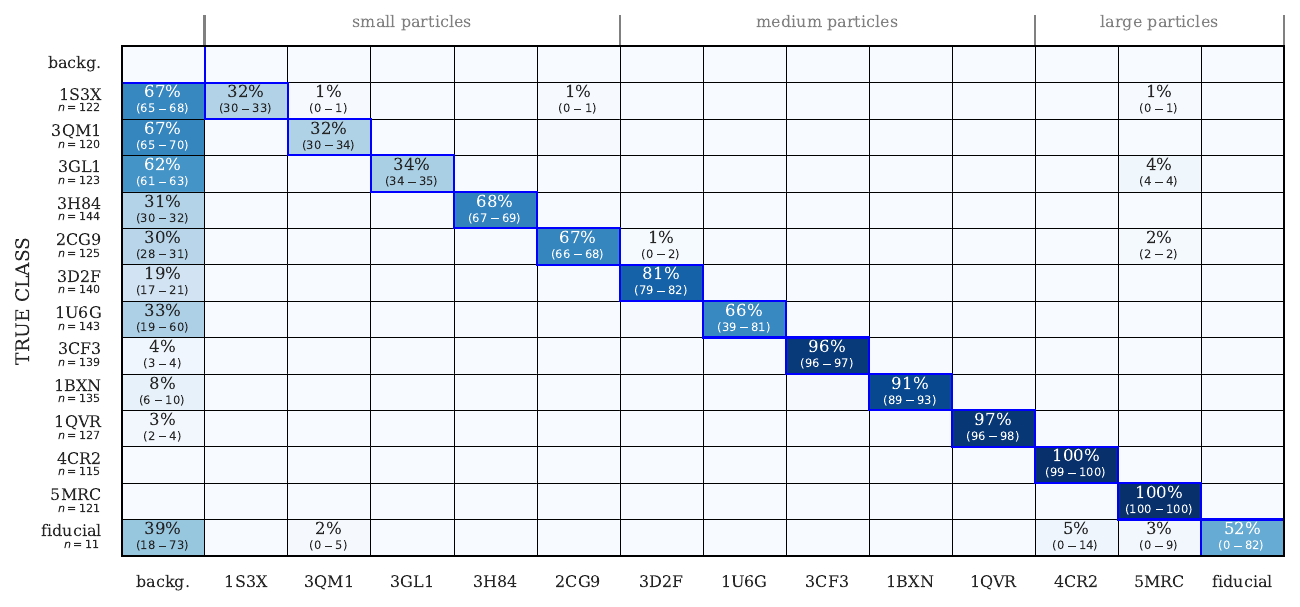} }}%
    \caption{Confusion matrices for {\sc baseline} (top) and \emph{noiseless} (bottom) experiments\ifcell , related to STAR Methods\fi.}
    \label{fig:confmats_baseline_noiseless}%
\end{figure}

\begin{figure}
    \centering
    \subfloat{{\label{fig:confusion_matrix_finetuned-3}
        \includegraphics[width=0.98\linewidth]{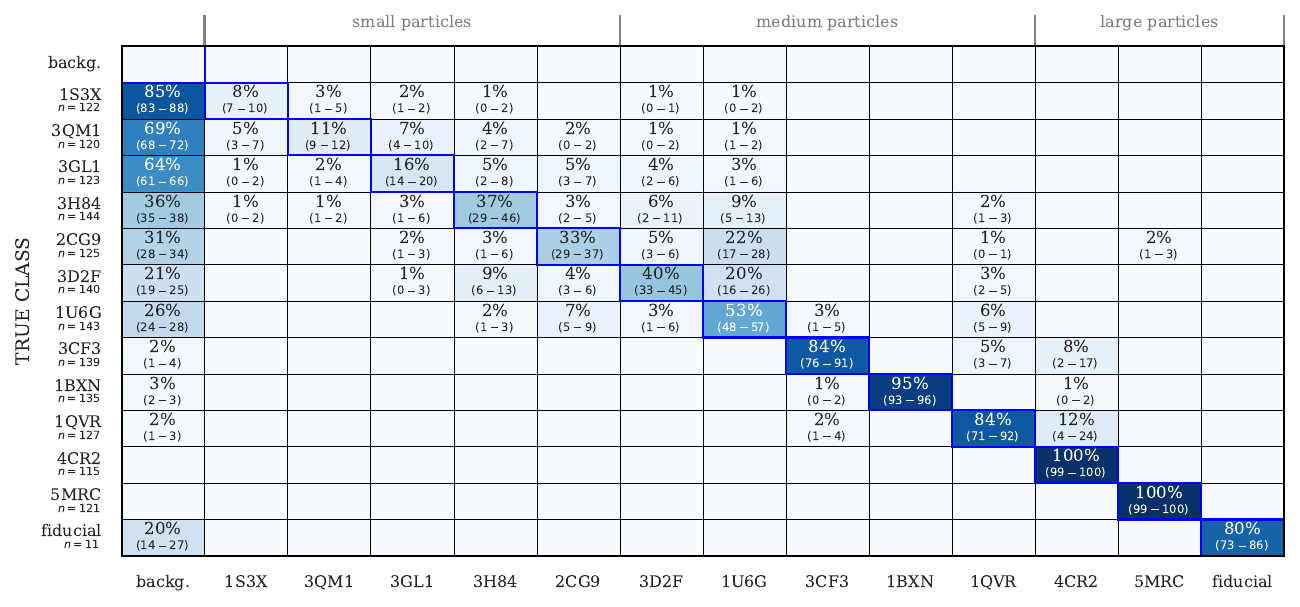} }}%
            \\
    \subfloat{{\label{fig:confusion_matrix_benchmark-3}
        \includegraphics[width=0.98\linewidth]{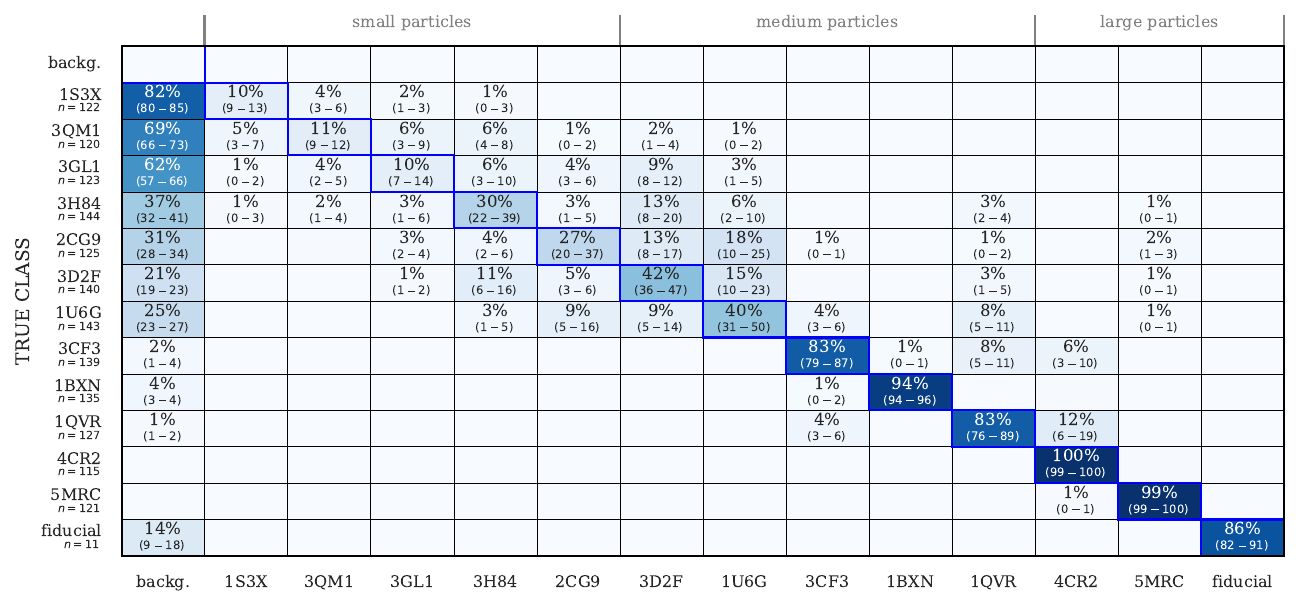} }}%
    \caption{Confusion matrices for {\sc finetuned-3} (top) and {\sc benchmark-3} (bottom) experiments\ifcell , related to STAR Methods\fi.}
    \label{fig:confmats_finetuned-3_benchmark-3}%
\end{figure}


\clearpage
\ifarxiv
\hlb{
\section{Profiling execution time and memory consumption}
\label{appendix:profiling}
}
\else 
\hlb{
\section{Profiling execution time and memory consumption, related to \nameref{subsec:nst}}
\label{appendix:profiling}
}
\fi

\hlb{
\paragraph{What hardware was used?} Profiling of FakET was done using an NVIDIA A100 40GB SXM4 GPU, a Dual AMD EPYC 7742 64-Core CPU with total 128 physical CPU cores and 256 threads, a DDR4-3200\,MHz memory, and Samsung PM983 NVMe read-intensive hard drives.
}

\hlb{
\paragraph{What profiling tools were used?} Measuring the RAM usage was done using the \texttt{memory-profiler v0.61.0} Python package, more precisely using the following command \texttt{mprof run -T 0.1} that logs the RAM usage of a process in mebibytes (MiB) with the time resolution of 0.1\,s. Measuring the VRAM usage in MiB was done using the \texttt{nvidia-ml-py3 v7.352.0} Python package, more precisely using repeated calls of \texttt{nvidia\_smi.nvmlDeviceGetMemoryInfo()} function every 0.05\,s. For each profiled scenario, we noted the total duration of the execution and the peak memory usage. In order to get consistent results and predictable memory usage, all of the measurements in \Cref{table:profiling} were done with the flags \texttt{torch.backends.cudnn.benchmark = False} and \texttt{torch.backends.cudnn.deterministic = True}, however shorter execution times can be achieved for the cost of increased memory footprint. The speed and VRAM usage reported in the main body of the article was measured with \texttt{torch.backends.cudnn.benchmark = True} and \texttt{torch.backends.cudnn.deterministic = False}.
}

\hlb{
\paragraph{What scenarios were profiled?} In order to provide the measurements for scenarios useful to practitioners, we chose to profile simulation of single micrographs as well as whole tilt-series of 61 projections. Both in two different sizes -- $1024\times1024$ (as in SHREC) and $\approx12\times$ larger $3500\times3500$ (usual-sized).}

\hlb{
\paragraph{What can be simulated in what time?} Using just 8\,CPUs and 6\,GiB of RAM, 10 SHREC-sized tomograms can be simulated in under 80 minutes. Using 32\,CPUs and the same amount of RAM, the same can be achieved in under 45 minutes. Assuming availability of 8 GPUs with 40GiB of VRAM, one can create a data set of 100 tilt-series of usual size in under 2h.}

\hlb{
\paragraph{Am I going to be able to run FakET on my machine?}
In case hardware resources are scarce, FakET can still be utilized to generate micrographs or tilt-series by trading off the execution time. For instance, with only $1\times$\,CPU and $4\,GiB$ of RAM, we were able to simulate a SHREC-like data set one micrograph at a time while the computation was still $22\times$\,faster than SHREC simulator in terms of wall-clock time. In another scenario, where the size of the simulated micrograph causes GPU out-of-memory errors, assuming abundance of RAM, one can revert to CPU-only computation again trading off execution time. E.g. in our case, it was no longer possible to simulate $61\times3500\times3500$ tilt-series using a GPU due to shortage of VRAM, but it was still possible to get on with the calculation using only CPUs and $\approx53\,GiB$ of RAM.
}

\hlb{
\paragraph{Practical usage tips:} •~When simulating on a CPU, using more cores than projections in a tilt-series is pointless, as projections are being processed in parallel each by one CPU. •~In case multiple GPUs are available, all can be utilized in parallel to simulate multiple samples. •~Parallel simulation of one tilt-series on multiple GPUs is not implemented but can be worked around by simply slicing the tilt-series into multiple batches of less tilts and simulating each batch on a different GPU. \texttt{faket.style\_transfer.cli} arguments \texttt{--seq\_start} and \texttt{--seq\_end} make this effortless. •~VRAM consumption can be further decreased by utilizing the style tensor of smaller size than the input tensors. We utilized this trick in order to be able to simulate $61\times3500\times3500$ tilt-series using a GPU with 40GiB of VRAM (see the fifth highlighted row in the~\Cref{table:profiling}) that would otherwise not suffice (see the last highlighted row in the aforementioned table).}

\begin{table*}[!ht]
\vskip 0.15in
\begin{center}
\begin{small}
\begin{sc}
\begin{tabular}{llrrrr}
\multicolumn{6}{c}{Micrographs (Single Tilts)} \\
\toprule
Input shapes & Style shape & Device & RAM [$GiB$] & VRAM [$GiB$] & Duration [$min$] \\ 
\midrule
\rowcolor{lightgray!40} 
$1\times1024\times1024$ & $1\times1024\times1024$ &  $1\times$GPU & $4.5$  & $7.2$      & $0.27$ \\ 
$1\times1024\times1024$ & $1\times1024\times1024$ &  $1\times$CPU & $3.9$  & -          & $0.67$ \\ 
\midrule
\rowcolor{lightgray!40}
$1\times3500\times3500$ & $1\times1024\times1024$ &  $1\times$GPU & $4.6$  & $38.6$     & $0.37$ \\ 
$1\times3500\times3500$ & $1\times1024\times1024$ &  $1\times$CPU & $33.5$ & -          & $5.52$ \\ 
\midrule
\rowcolor{lightgray!40}
$1\times3500\times3500$ & $1\times3500\times3500$ &  $1\times$GPU & \multicolumn{3}{r}{DOES NOT FIT ON NVIDIA A100-40GB GPU} \\
$1\times3500\times3500$ & $1\times3500\times3500$ &  $1\times$CPU & $38.8$ & -          & $7.83$ \\ 
\bottomrule
\\
\\
\multicolumn{6}{c}{Tilt-series} \\
\toprule
Input shapes & Style shape & Device & RAM [$GiB$] & VRAM [$GiB$] & Duration [$min$] \\
\midrule
\rowcolor{lightgray!40}
$61\times1024\times1024$ & $61\times1024\times1024$ &  $1\times$GPU  & $6.3$ & $7.2$   & $1.50$ \\ 
$61\times1024\times1024$ & $61\times1024\times1024$ &  $64\times$CPU & $5.7$ & -       & $4.15$ \\ 
$61\times1024\times1024$ & $61\times1024\times1024$ &  $32\times$CPU & $5.2$ & -       & $4.38$ \\ 
$61\times1024\times1024$ & $61\times1024\times1024$ &  $16\times$CPU & $5.4$ & -       & $6.03$ \\ 
$61\times1024\times1024$ & $61\times1024\times1024$ &  $8\times$CPU  & $5.3$ & -       & $7.95$ \\ 
\midrule
\rowcolor{lightgray!40}
$61\times3500\times3500$ & $61\times1024\times1024$ &  $1\times$GPU  & $21.3$ & $38.6$ & $8.95$ \\ 
$61\times3500\times3500$ & $61\times1024\times1024$ &  $64\times$CPU & $42.7$ & -      & $40.38$ \\ 
$61\times3500\times3500$ & $61\times1024\times1024$ &  $32\times$CPU & $42.4$ & -      & $42.80$ \\ 
$61\times3500\times3500$ & $61\times1024\times1024$ &  $16\times$CPU & $42.3$ & -      & $60.03$ \\ 
$61\times3500\times3500$ & $61\times1024\times1024$ &  $8\times$CPU  & $42.3$ & -      & $77.45$ \\ 
\midrule
\rowcolor{lightgray!40}
$61\times3500\times3500$ & $61\times3500\times3500$ &  $1\times$GPU  & \multicolumn{3}{r}{DOES NOT FIT ON NVIDIA A100-40GB GPU} \\
$61\times3500\times3500$ & $61\times3500\times3500$ &  $64\times$CPU & $52.9$ & -      & $51.58$ \\ 
\bottomrule
\end{tabular}
\end{sc}
\end{small}
\end{center}
\vskip -0.1in
\caption{\hlb{Measurements of execution time and peak RAM \& VRAM consumption during simulation of each scenario\ifcell , related to \nameref{subsec:nst}\fi.  
Notation: Input shapes -- the shape of the ``init'' and ``content'' tensors; Style shape -- the shape of the ``style'' tensor. The rows where GPU was utilized are highlighted.
}}
\label{table:profiling}
\end{table*}
\clearpage
\renewcommand{\thefigure}{S\arabic{figure}}
\renewcommand{\thetable}{S\arabic{table}}
\captionsetup[table]{position=bottom, name=Table} 
\captionsetup[figure]{position=bottom, name=Figure} 
\setcounter{figure}{0} 
\setcounter{table}{0}  

\pagestyle{empty}

\ifcell 
\pagestyle{customfooter}

\clearpage
\else 
\clearpage
\section*{Supplementary Information}
\fi

\begin{table*}[!h]
\begin{center}
\begin{tabular}{p{1.0cm}p{1.9cm}rrrrrrrrrrrrrrr}
& & & & && & & && & & && & & \\
  \rotatebox[origin=l]{90}{model} & 
  \rotatebox[origin=l]{90}{modality} &  
  \rotatebox[origin=l]{90}{MAE mean} &  
  \rotatebox[origin=l]{90}{MAE $10^{th}$} &  
  \rotatebox[origin=l]{90}{MAE $90^{th}$} &  
  &
  \rotatebox[origin=l]{90}{MSE mean} &  
  \rotatebox[origin=l]{90}{MSE $10^{th}$} &  
  \rotatebox[origin=l]{90}{MSE $90^{th}$} &  
  &
  \rotatebox[origin=l]{90}{PSNR mean} &  
  \rotatebox[origin=l]{90}{PSNR $10^{th}$} &  
  \rotatebox[origin=l]{90}{PSNR $90^{th}$} &  
  &
  \rotatebox[origin=l]{90}{SSIM mean} &  
  \rotatebox[origin=l]{90}{SSIM $10^{th}$} &  
  \rotatebox[origin=l]{90}{SSIM $90^{th}$} \\
\midrule \noalign{\vspace{0.3em}}
\multirow[c]{3}{*}{all}
 & faket & 0.37 & 0.32 & 0.45 && 0.15 & 0.11 & 0.21 && 8.40 & 6.80 & 9.57 && 0.25 & 0.17 & 0.38 \\
 & noisy & 0.38 & 0.35 & 0.42 && 0.15 & 0.13 & 0.18 && 8.16 & 7.44 & 8.86 && 0.19 & 0.17 & 0.21 \\
 & shrec & 0.53 & 0.47 & 0.61 && 0.29 & 0.23 & 0.38 && 5.45 & 4.22 & 6.44 && 0.17 & 0.13 & 0.21 \\ \noalign{\vspace{0.3em}}
\midrule \noalign{\vspace{0.3em}}
\multirow[c]{3}{*}{0} 
 & faket & 0.38 & 0.34 & 0.42 && 0.16 & 0.12 & 0.18 && 8.14 & 7.38 & 9.12 && 0.19 & 0.17 & 0.20 \\
 & noisy & 0.41 & 0.39 & 0.43 && 0.18 & 0.16 & 0.19 && 7.47 & 7.20 & 7.97 && 0.18 & 0.16 & 0.19 \\
 & shrec & 0.60 & 0.53 & 0.64 && 0.37 & 0.29 & 0.42 && 4.34 & 3.79 & 5.37 && 0.13 & 0.13 & 0.14 \\ \noalign{\vspace{0.3em}}
\multirow[c]{3}{*}{1} 
 & faket & 0.36 & 0.35 & 0.39 && 0.14 & 0.13 & 0.16 && 8.56 & 8.08 & 8.87 && 0.18 & 0.16 & 0.22 \\
 & noisy & 0.37 & 0.34 & 0.38 && 0.14 & 0.12 & 0.15 && 8.54 & 8.25 & 9.07 && 0.20 & 0.18 & 0.21 \\
 & shrec & 0.51 & 0.47 & 0.55 && 0.27 & 0.23 & 0.31 && 5.67 & 5.12 & 6.41 && 0.17 & 0.15 & 0.17 \\ \noalign{\vspace{0.3em}}
\multirow[c]{3}{*}{2} 
 & faket & 0.33 & 0.30 & 0.35 && 0.11 & 0.10 & 0.12 && 9.47 & 9.07 & 10.20 && 0.38 & 0.36 & 0.41 \\
 & noisy & 0.38 & 0.36 & 0.39 && 0.15 & 0.13 & 0.16 && 8.22 & 7.94 & 8.75 && 0.19 & 0.18 & 0.20 \\
 & shrec & 0.53 & 0.52 & 0.55 && 0.29 & 0.28 & 0.31 && 5.37 & 5.02 & 5.58 && 0.13 & 0.11 & 0.16 \\ \noalign{\vspace{0.3em}}
\multirow[c]{3}{*}{3} 
 & faket & 0.34 & 0.31 & 0.36 && 0.12 & 0.10 & 0.13 && 9.19 & 8.79 & 9.89 && 0.36 & 0.33 & 0.38 \\
 & noisy & 0.39 & 0.37 & 0.41 && 0.16 & 0.14 & 0.17 && 7.87 & 7.59 & 8.41 && 0.17 & 0.15 & 0.17 \\
 & shrec & 0.48 & 0.44 & 0.50 && 0.24 & 0.20 & 0.26 && 6.28 & 5.88 & 6.98 && 0.20 & 0.19 & 0.21 \\ \noalign{\vspace{0.3em}}
\multirow[c]{3}{*}{4} 
 & faket & 0.38 & 0.33 & 0.45 && 0.15 & 0.12 & 0.21 && 8.29 & 6.88 & 9.22 && 0.20 & 0.17 & 0.22 \\
 & noisy & 0.37 & 0.35 & 0.39 && 0.14 & 0.13 & 0.15 && 8.40 & 8.13 & 8.92 && 0.20 & 0.18 & 0.21 \\
 & shrec & 0.48 & 0.44 & 0.50 && 0.24 & 0.20 & 0.26 && 6.27 & 5.87 & 6.97 && 0.21 & 0.20 & 0.22 \\ \noalign{\vspace{0.3em}}
\multirow[c]{3}{*}{5} 
 & faket & 0.36 & 0.33 & 0.38 && 0.13 & 0.11 & 0.15 && 8.73 & 8.34 & 9.45 && 0.34 & 0.33 & 0.37 \\
 & noisy & 0.41 & 0.38 & 0.42 && 0.17 & 0.15 & 0.18 && 7.62 & 7.36 & 8.13 && 0.18 & 0.17 & 0.19 \\
 & shrec & 0.52 & 0.48 & 0.60 && 0.29 & 0.24 & 0.37 && 5.49 & 4.36 & 6.23 && 0.14 & 0.13 & 0.15 \\ \noalign{\vspace{0.3em}}
\multirow[c]{3}{*}{6} 
 & faket & 0.38 & 0.32 & 0.41 && 0.15 & 0.11 & 0.17 && 8.31 & 7.59 & 9.56 && 0.24 & 0.22 & 0.26 \\
 & noisy & 0.35 & 0.33 & 0.36 && 0.13 & 0.11 & 0.14 && 8.93 & 8.65 & 9.47 && 0.21 & 0.19 & 0.22 \\
 & shrec & 0.50 & 0.46 & 0.52 && 0.26 & 0.22 & 0.28 && 5.93 & 5.52 & 6.65 && 0.19 & 0.18 & 0.20 \\ \noalign{\vspace{0.3em}}
\multirow[c]{3}{*}{7} 
 & faket & 0.43 & 0.39 & 0.50 && 0.19 & 0.16 & 0.25 && 7.23 & 5.99 & 8.05 && 0.19 & 0.17 & 0.22 \\
 & noisy & 0.39 & 0.37 & 0.41 && 0.16 & 0.14 & 0.17 && 7.93 & 7.66 & 8.45 && 0.19 & 0.17 & 0.20 \\
 & shrec & 0.57 & 0.51 & 0.60 && 0.33 & 0.26 & 0.37 && 4.89 & 4.37 & 5.78 && 0.17 & 0.17 & 0.18 \\ \noalign{\vspace{0.3em}}
\multirow[c]{3}{*}{8} 
 & faket & 0.42 & 0.34 & 0.47 && 0.19 & 0.13 & 0.22 && 7.36 & 6.49 & 8.99 && 0.19 & 0.18 & 0.19 \\
 & noisy & 0.38 & 0.36 & 0.39 && 0.15 & 0.13 & 0.16 && 8.20 & 7.93 & 8.72 && 0.20 & 0.18 & 0.21 \\
 & shrec & 0.52 & 0.48 & 0.59 && 0.28 & 0.24 & 0.35 && 5.53 & 4.51 & 6.20 && 0.17 & 0.16 & 0.19 \\ \noalign{\vspace{0.3em}}
\multirow[c]{3}{*}{9} 
 & faket & 0.36 & 0.31 & 0.39 && 0.14 & 0.11 & 0.16 && 8.70 & 7.89 & 9.76 && 0.20 & 0.19 & 0.21 \\
 & noisy & 0.37 & 0.35 & 0.38 && 0.15 & 0.13 & 0.15 && 8.38 & 8.10 & 8.92 && 0.19 & 0.17 & 0.20 \\
 & shrec & 0.57 & 0.50 & 0.61 && 0.34 & 0.26 & 0.38 && 4.72 & 4.15 & 5.78 && 0.14 & 0.13 & 0.15 \\ \noalign{\vspace{0.3em}}
\bottomrule
\end{tabular}
\end{center}
\vspace{-0.6em}
\caption{Standard metrics\hlb{, related to \nameref{sec:evaluation}}. Despite evaluating the quality of our simulations directly on downstream tasks (particle localization and classification), we also compute standard metrics. Mean Absolute Error (MAE), Mean Squared Error (MSE), Peak Signal-to-Noise Ratio (PSNR), Structural Similarity Index (SSIM) metrics for simulated tilt-series are computed in contrast to their noiseless counterparts. Since SSIM assumes non-negative images with a well-defined range of values, we normalize each tilt-series to interval~$[0, 1]$. Then we compute the MAE, MSE, PSNR, and SSIM per tilt and provide the mean value of each metric per tilt-series along with its $10^{th}$ and $90^{th}$ percentiles.}
\label{table:standardmetrics}
\end{table*}

\begin{figure*}
\begin{center}
\includegraphics[width=0.98\textwidth]{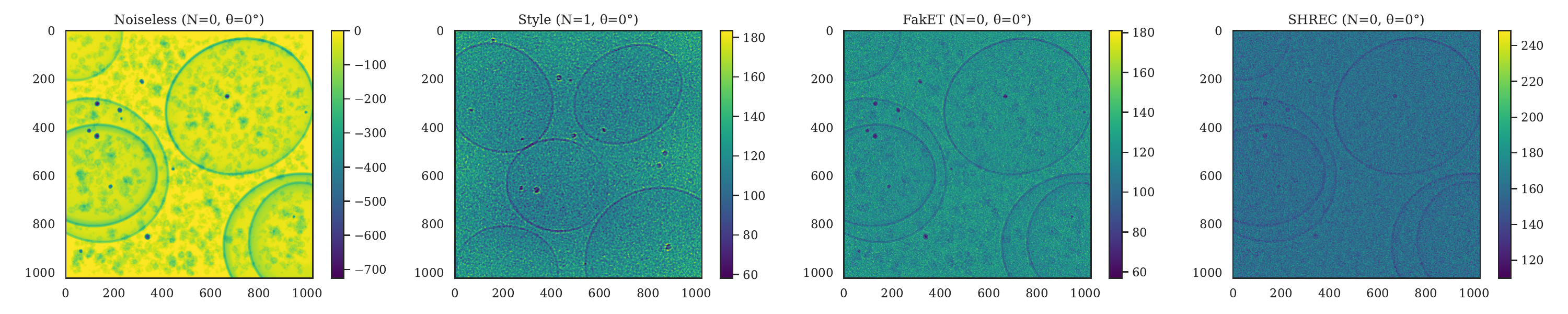}
\includegraphics[width=0.98\textwidth]{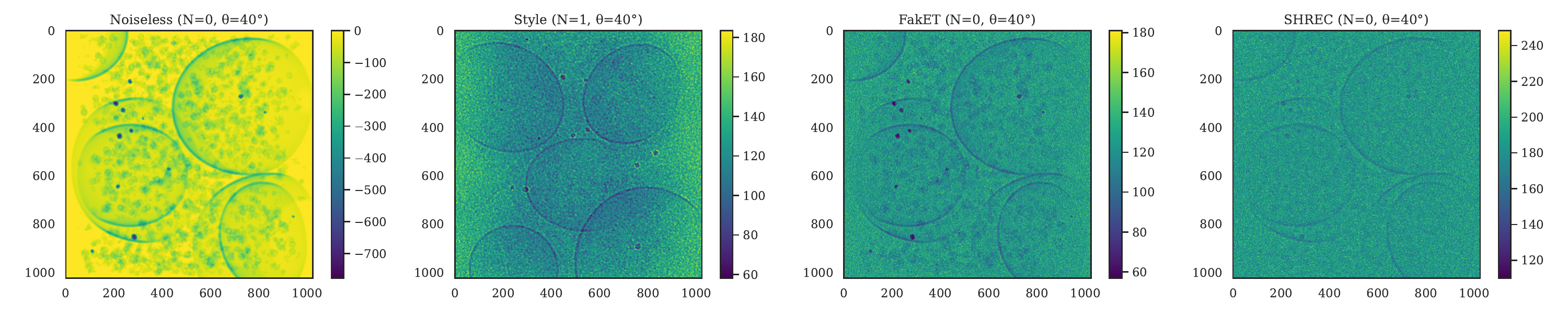}
\includegraphics[width=0.98\textwidth]{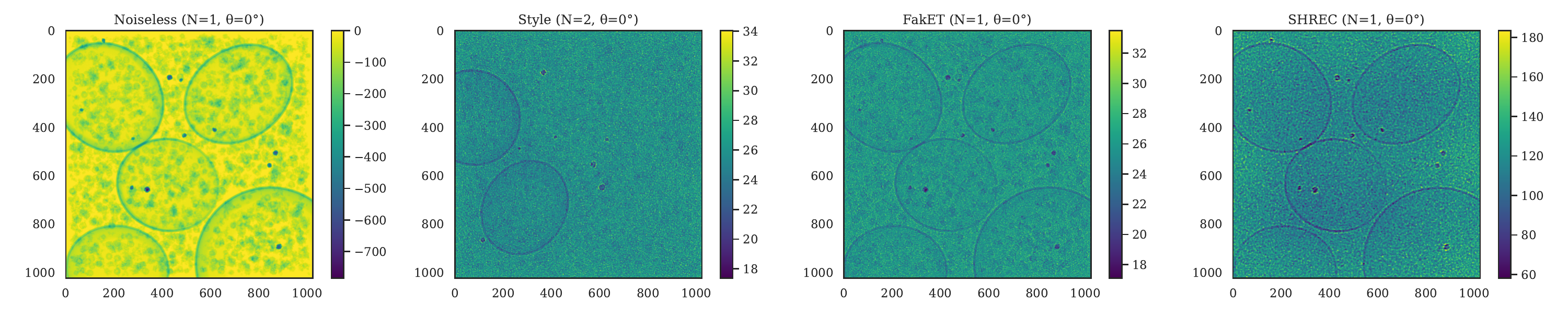}
\includegraphics[width=0.98\textwidth]{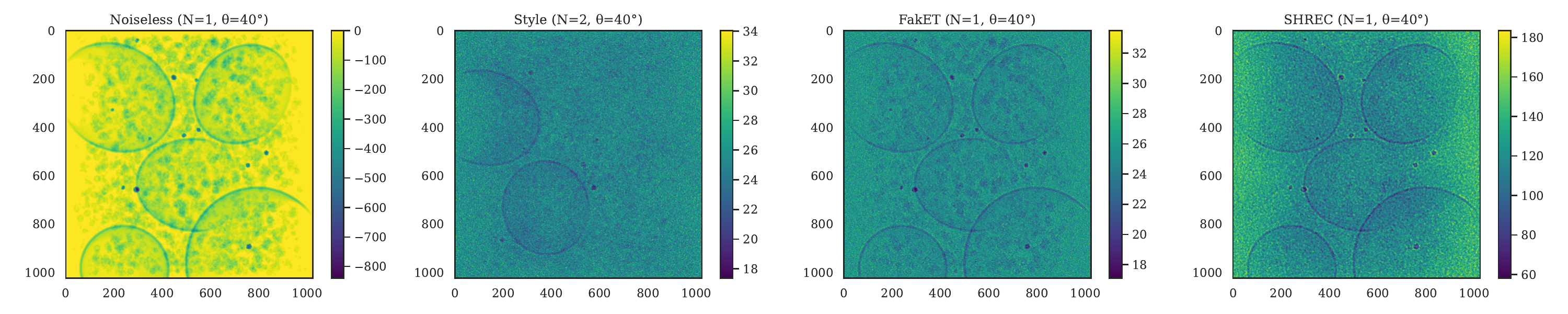}
\includegraphics[width=0.98\textwidth]{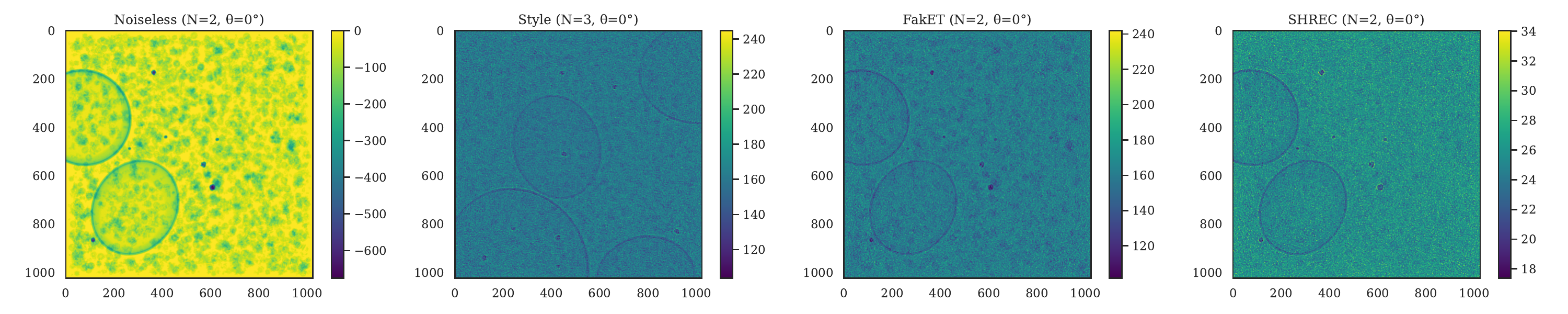}
\includegraphics[width=0.98\textwidth]{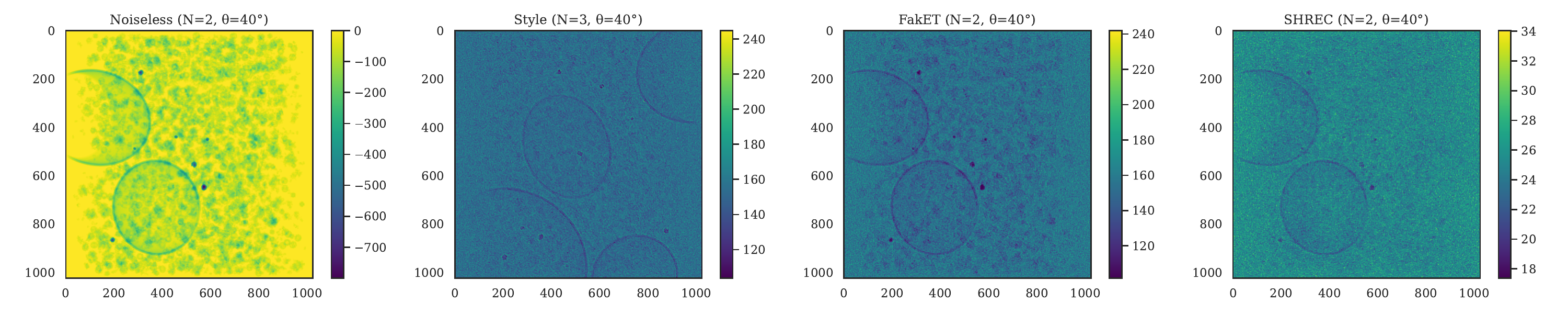}
\end{center}
\caption{Visualization of the inputs (Noiseless, Style), output (FakET) and ground truth (SHREC) projections of the first 3 training tomograms at 0° and 40°\ifcell \hlb{, related to STAR Methods}\fi. Style is always taken from the $(N+1)^{th}$ training tomogram. To minimize the impact of random extreme values (outliers) on contrast, we visualize the Style, FakET, and SHREC projections within the $1^{st}$ and $99^{th}$ percentiles of the pixel intensity. The percentiles are computed based on the whole tilt-series, not just one tilt. We chose the viridis over greyscale colormap for its perceptual uniformity, accessibility for color vision deficiencies, and enhanced detail visibility. Colorbar indicates simulated intensities in arbitrary units.}
\label{supfig:inputs1}
\end{figure*}

\begin{figure*}
\begin{center}
\includegraphics[width=0.98\textwidth]{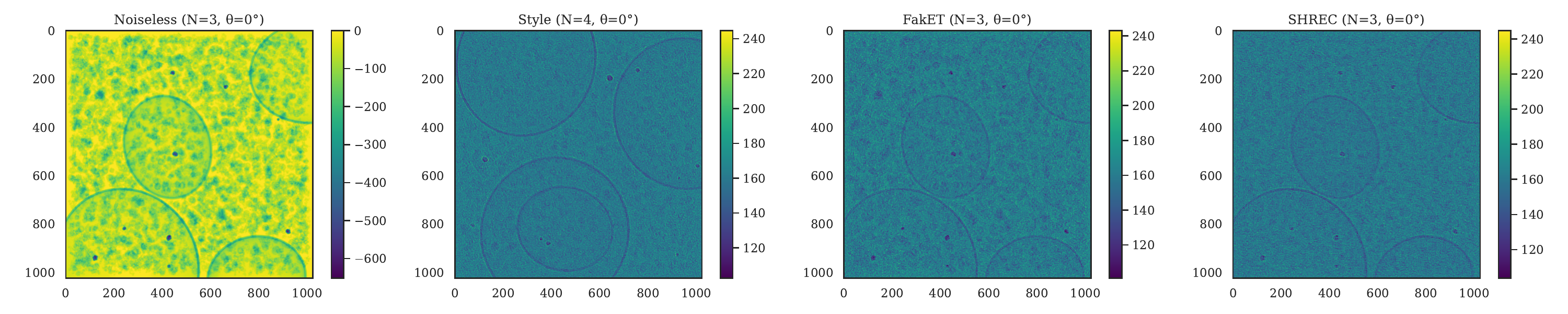}
\includegraphics[width=0.98\textwidth]{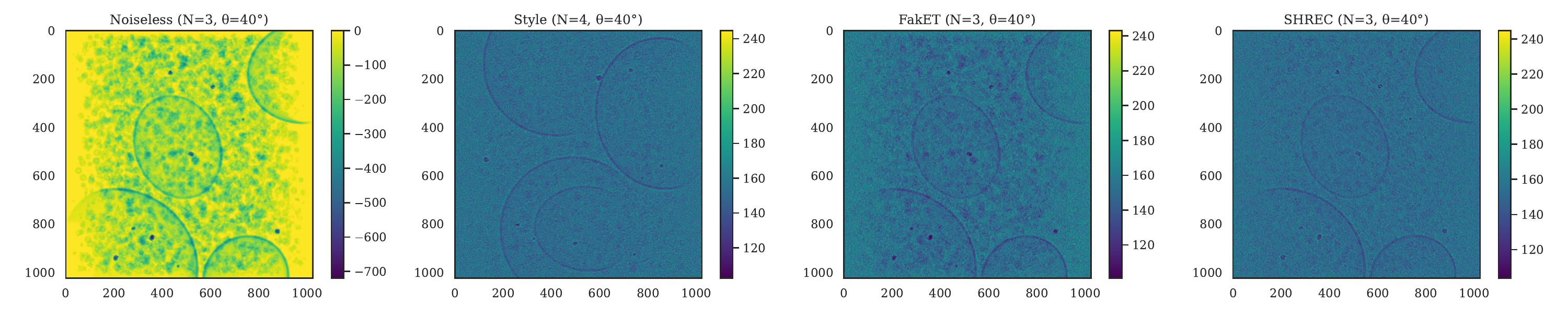}
\includegraphics[width=0.98\textwidth]{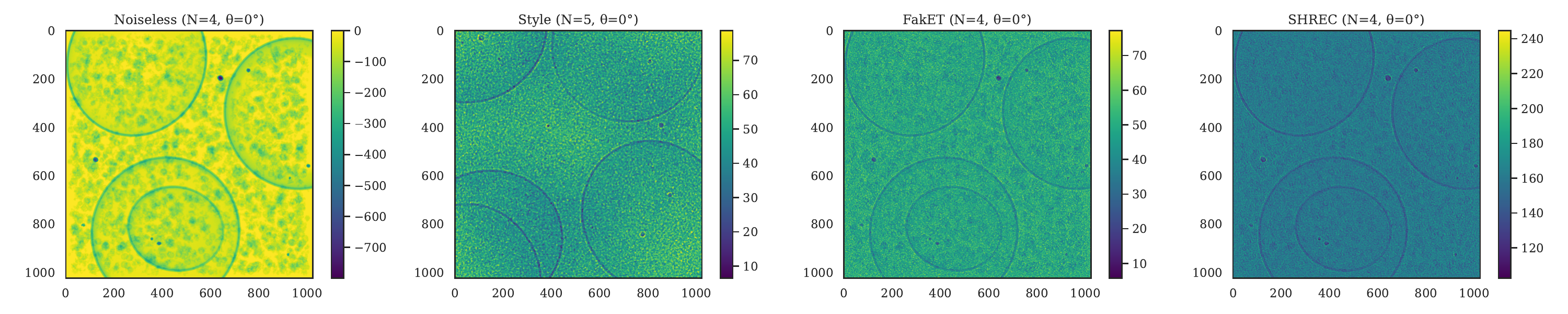}
\includegraphics[width=0.98\textwidth]{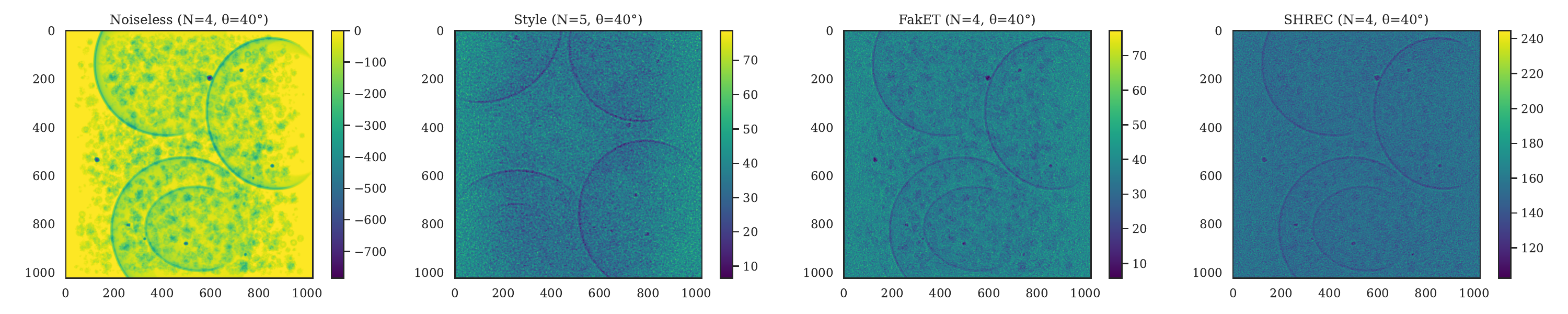}
\includegraphics[width=0.98\textwidth]{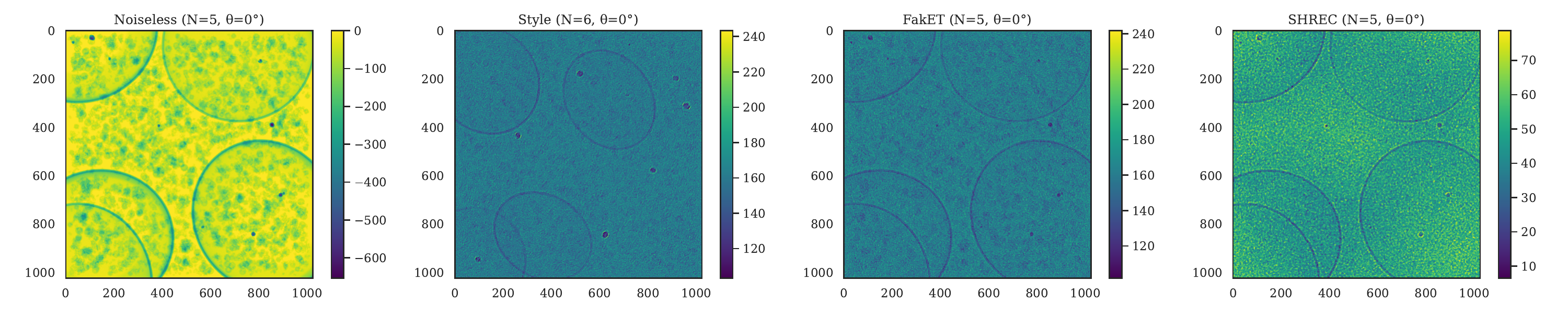}
\includegraphics[width=0.98\textwidth]{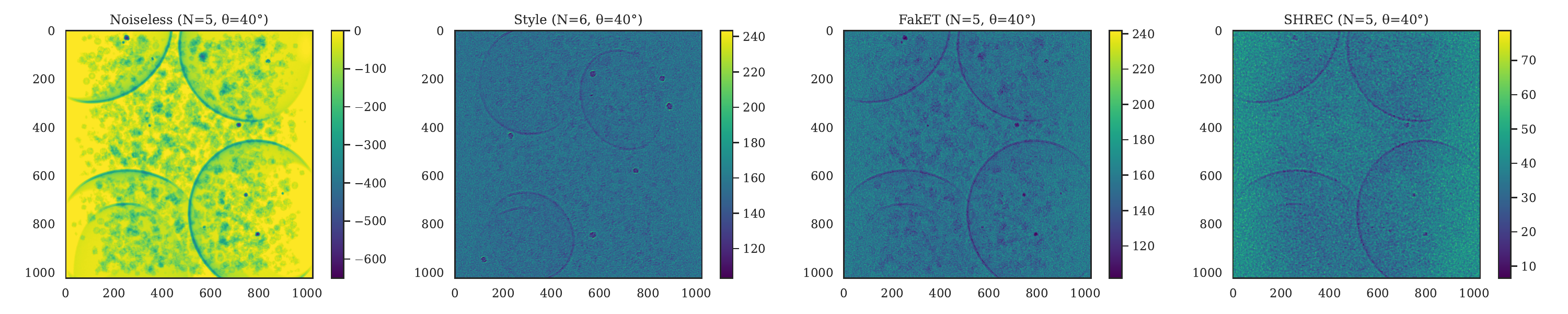}
\end{center}
\caption{Visualization of the inputs (Noiseless, Style), output (FakET) and ground truth (SHREC) projections of the $4^{th}$ - $6^{th}$ training tomograms at 0° and 40°\ifcell \hlb{, related to STAR Methods}\fi. Style is always taken from the $(N+1)^{th}$ training tomogram. To minimize the impact of random extreme values (outliers) on contrast, we visualize the Style, FakET, and SHREC projections within the $1^{st}$ and $99^{th}$ percentiles of the pixel intensity. The percentiles are computed based on the whole tilt-series, not just one tilt. We chose the viridis over greyscale colormap for its perceptual uniformity, accessibility for color vision deficiencies, and enhanced detail visibility. Colorbar indicates simulated intensities in arbitrary units.}
\label{supfig:inputs2}
\end{figure*}

\begin{figure*}
\begin{center}
\includegraphics[width=0.98\textwidth]{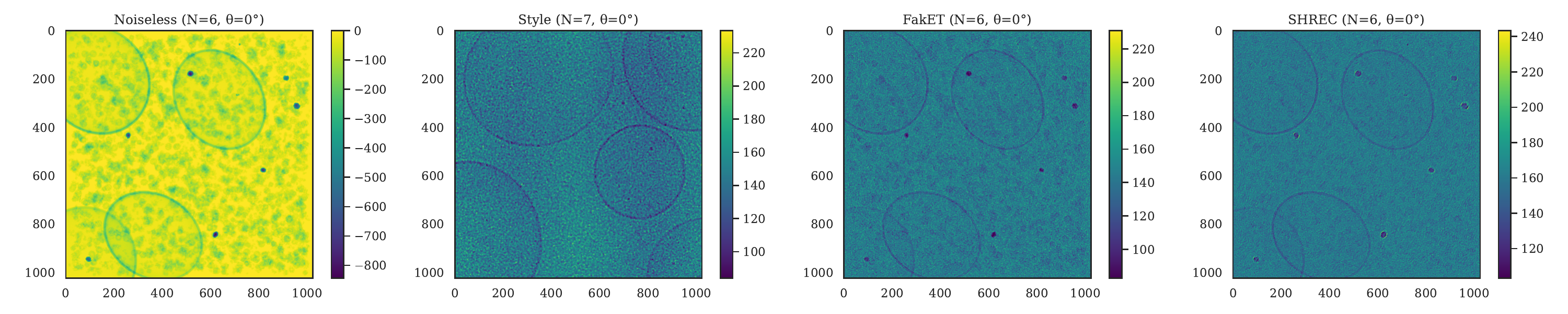}
\includegraphics[width=0.98\textwidth]{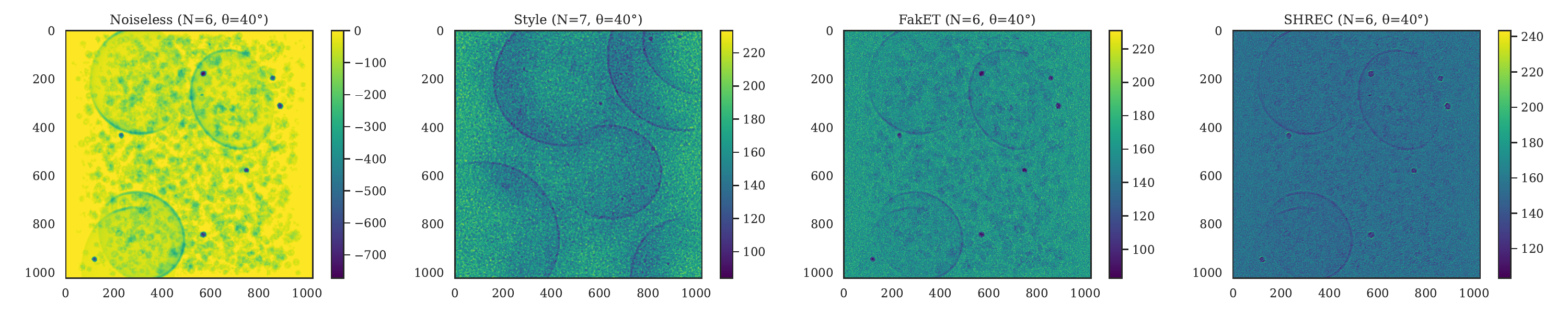}
\includegraphics[width=0.98\textwidth]{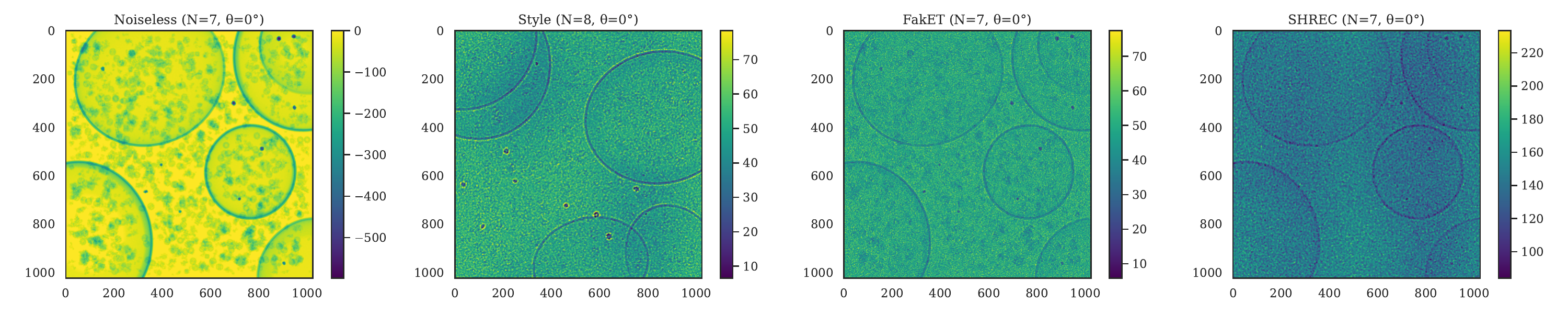}
\includegraphics[width=0.98\textwidth]{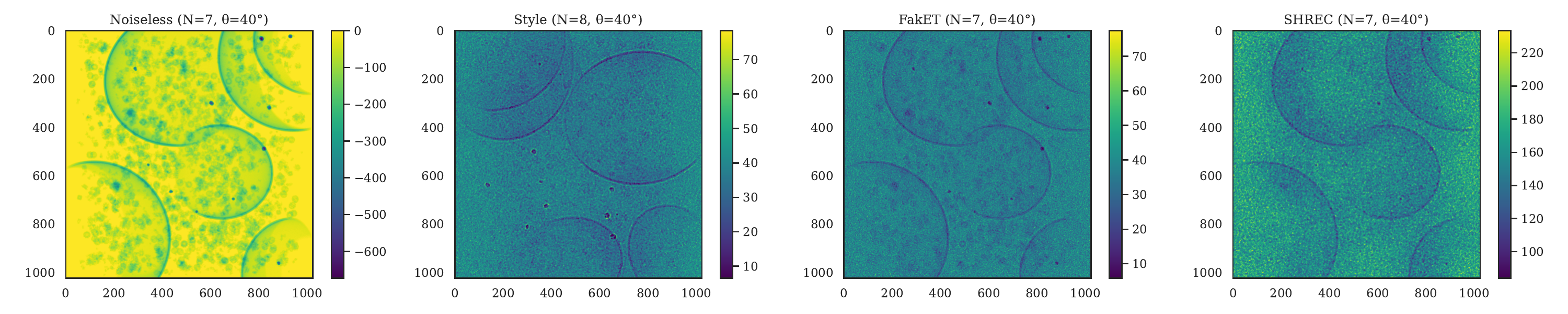}
\includegraphics[width=0.98\textwidth]{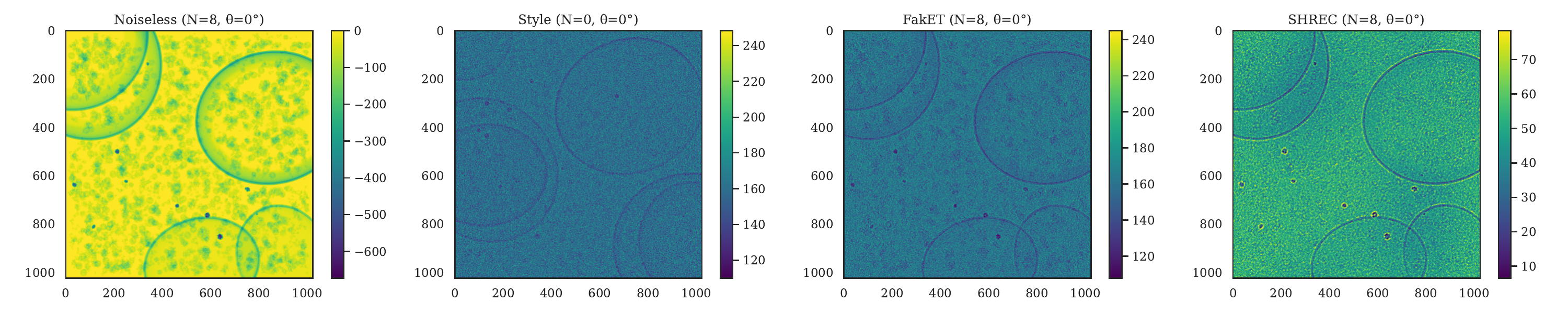}
\includegraphics[width=0.98\textwidth]{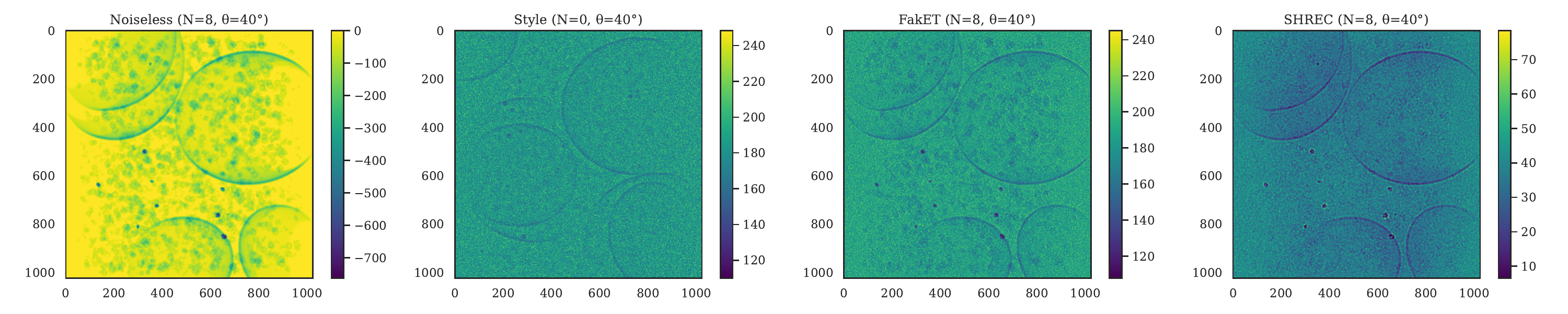}
\end{center}
\caption{Visualization of the inputs (Noiseless, Style), output (FakET) and ground truth (SHREC) projections of the last 3 training tomograms at 0° and 40°\ifcell \hlb{, related to STAR Methods}\fi. Style is always taken from the $(N+1)^{th}$ training tomogram. To minimize the impact of random extreme values (outliers) on contrast, we visualize the Style, FakET, and SHREC projections within the $1^{st}$ and $99^{th}$ percentiles of the pixel intensity. The percentiles are computed based on the whole tilt-series, not just one tilt. We chose the viridis over greyscale colormap for its perceptual uniformity, accessibility for color vision deficiencies, and enhanced detail visibility. Colorbar indicates simulated intensities in arbitrary units.}
\label{supfig:inputs3}
\end{figure*}
\else

\renewcommand\hl[1]{{#1}} 
\renewcommand\hlb[1]{{#1}} 

\clearpage
\onecolumn

\renewcommand{\thesection}{Data S\arabic{section}}
\setcounter{section}{0} 
\crefname{section}{Data}{Data}
\Crefname{section}{Data}{Data}
\makeatletter
\renewcommand{\theHsection}{Data.S\arabic{section}}
\makeatother

\fi






\end{document}